\documentclass{article}

    \usepackage[preprint]{neurips_2024}

\usepackage[utf8]{inputenc} 
\usepackage[T1]{fontenc}    
\usepackage{hyperref}       
\usepackage{url}            
\usepackage{booktabs}       
\usepackage{amsfonts}       
\usepackage{nicefrac}       
\usepackage{microtype}      
\usepackage{xcolor}         
\usepackage{microtype}
\usepackage{graphicx}
\usepackage{subcaption}
\usepackage{booktabs}
\usepackage{hyperref}
\usepackage{amsmath}
\usepackage{amssymb}
\usepackage{mathtools}
\usepackage{amsthm}
\usepackage[capitalize,noabbrev]{cleveref}
\usepackage{colortbl}

\theoremstyle{plain}

\theoremstyle{definition}

\theoremstyle{remark}

\newcommand{\RMS}{\mathrm{RMS}}
\newcommand{\LAR}{\mathrm{LAR}}
\newcommand{\pred}{\text{pred}}

\newcommand{\corr}{\text{corr}}

\DeclareMathOperator{\rank}{rank}

\title{A Training-Time Diagnostic for Generalization via the Log-Alignment Ratio}

 \author{%
    Ali Shehper\thanks{Work done at Essential AI.} \\
    \texttt{ali.shehper1@gmail.com} \\
    \And
    Ashish Vaswani\footnotemark[1] \\
    \texttt{ashish.vaswani@gmail.com} \\
  }

\begin{document}

\maketitle

\begin{abstract}
We study the log-alignment ratio (LAR), a measure of parameter–activation alignment,     
 introduced in parameterization theory. We reformulate it as the overlap between a weight spectrum $p$ 
 of the normalized squared singular values of a matrix and an activation spectrum $q$ of the normalized   
 squared projections of inputs onto its singular directions. We show that unembedding LAR tracks the transition between 
memorization and generalization in two different settings by capturing the spread of $p$ and $q$ during training. 
In grokking, LAR predicts the effective dimension of the learned function: $k \approx n^{2(1-\LAR)}$, where $n$ is the input dimension of the matrix. 
In 3B-parameter language model pre-training, its deviation from a non-overfitting baseline tracks the generalization gap, and its rate of decline increases as overfitting approaches. 
LAR is computable from quantities available during the forward pass with negligible computational overhead, and requires no held-out validation data.
\end{abstract}

\section{Introduction}

Large-scale pre-training has driven the most visible recent advances in neural network capability \citep{kaplan2020scaling,hoffmann2022chinchilla}, and as model and dataset sizes continue to grow, the cost of every training run grows with them. This has made the study of training dynamics, and the design of cheap diagnostic tools for monitoring them, unusually valuable: a metric that can flag problems such as overfitting, instability, or saturation from training-time quantities alone, without requiring expensive held-out evaluation, can save compute and time for practitioners.

A complementary line of work on neural network \emph{parameterizations}, which prescribe how initialization, parameter multipliers, and learning rates should scale with width and depth, offers a principled approach to transferring hyperparameters from small to large models. \citet{yang2024tensor} proposed the maximal update parametrization ($\mu P$), which enables hyperparameter transfer under the assumption that updates at the readout layer are fully aligned with the incoming activations. \citet{everett2024scaling} showed that this alignment assumption is stronger than necessary and, by relaxing it, obtained a standard-parameterization recipe that consistently outperformed $\mu P$ across tens of thousands of experiments at scales up to 26.8B parameters. As part of this analysis they introduced the \emph{log-alignment ratio} (LAR) as an empirical measure of how strongly a layer's parameter matrix aligns with its input activations during training.

While \citet{everett2024scaling} used LAR as a proxy for parameter-activation alignment,
we discover in this paper that the metric carries a second, previously-unexplored meaning: namely, it measures how diffuse or concentrated the layer's activations and weight spectra are. 
More precisely, we show that LAR measures the overlap between a weight spectrum $p$ and an activation spectrum $q$: it is large when both
spectra are concentrated on the same small set of directions, and small when either is diffuse or when the two are  misaligned.

This insight makes LAR a natural candidate to measure \emph{generalization} --- the extent to which a model generalizes beyond its training data rather than overfitting to it. Empirically, we observe that well-generalizing networks tend to concentrate their computation on a small number of directions, while overfitting networks tend to spread it across many. Most previously-studied metrics for this behavior cover either the weight spectrum or the activation spectrum, but not both, and are computationally expensive as they require full SVD of the weight matrix or a PCA of the activations. In contrast, LAR looks at both weights and activations and is computable from quantities already available in the model's forward pass, with negligible computational overhead.

We study the relationship between generalization and the LAR of a model's unembedding matrix in two different settings.
First, on small algorithmic tasks exhibiting grokking \citep{power2022grokking}, we find that optimization 
concentrates $p$ and $q$ on a shared set of directions, which causes LAR to grow. As the number $k$ of directions becomes small, grokking may occur, depending on factors such as task difficulty and training dataset size. 
LAR predicts the dimensionality of the learned function, $k \approx n^{2(1-\LAR)}$, which we compare against $k_{95}$ --- the number of principal components explaining 95\% of the 
unembedding input activation variance.  
We find close agreement for final checkpoints across all experiments, including 12 grokking tasks and a sweep over training-set fractions.

Second, in 3B-parameter language model pre-training, we find
that unembedding LAR decreases slowly and stabilizes when the model generalizes, 
but declines sharply when overfitting is approaching.
The deviation of LAR from a non-overfitting baseline also closely tracks the generalization gap. The accelerating rate of decline in LAR prior to overfitting could enable evaluation-free methods to train generalizing models --- a possibility we leave open for future investigation.

Across both the grokking and large-scale settings, we discover a consistent picture: lower LAR corresponds to more diffuse weight and activation distributions, which coincides with memorization, while higher LAR corresponds to more concentrated distributions and generalization. 
In each case, we observe that the correlation between $p$ and $q$ remains high throughout training, so LAR effectively tracks how far the weight and activation distributions are from uniform.

\paragraph{Contributions.} Our main contributions are as follows.
\begin{itemize}
  \item \textbf{Distributional reformulation of LAR.} We show that LAR can be written as the log-overlap $\sum_i p_i q_i$ 
  between a weight spectrum $p$ and an activation spectrum $q$, together with equivalent covariance and correlation forms 
  that make the metric's behavior interpretable in terms of elementary statistical quantities.
  \item \textbf{Effective dimension via LAR.} We show that when $W$ concentrates its energy on $k$ singular directions and 
  activations lie along the same directions, $\LAR = 1 - \tfrac{1}{2}\log_n k$. 
  We verify this empirically in the grokking setting by comparing the LAR-predicted dimension to $k_{95}$ on 12 algorithmic tasks.
  \item \textbf{LAR as an overfitting diagnostic at scale.} On 3B-parameter Gemma-2 models trained with Adam and Muon, 
  we show that unembedding LAR stabilizes in the non-overfitting regime and declines sharply under overfitting, 
  and that its deviation from a non-overfitting baseline closely tracks the generalization gap.

  \item \textbf{LAR as a proxy for spread in $p$ and $q$.} We show that in practice, optimization drives $p$ and $q$ to be highly correlated, making $\LAR$ a proxy for the variance of $p$ and $q$. In pre-training experiments, the top singular value of the unembedding matrix is much larger than the rest and is the primary driver of the spread in $p$. This makes the contribution from the top singular direction, $L_1 = 1 + \tfrac{1}{2}\log_n(p_1 q_1)$, an alternative proxy for the pre-overfit signature of overfitting.

\end{itemize}

\section{Related Work}
\paragraph{Spectral measures of generalization.}

Several lines of work have connected spectral properties of weight matrices to generalization.
\citet{bartlett2017spectrally} and \citet{neyshabur2018pac} derived margin-based generalization
bounds in terms of products of spectral norms, and \citet{arora2018stronger} gave tighter
compression-based bounds using noise stability of trained networks. 
\citet{jiang2020fantastic} later
compared dozens of such measures in a large-scale empirical study.
Empirically, \citet{martin2021implicit} propose that the spectral density of weight matrices develops
heavy tails as training progresses, and use this as a data-free signal for generalization and early
stopping.
\citet{yunis2024spectral} show that the normalized effective rank of
weight matrices decreases broadly during training, and that the onset of low-rank weights coincides
with the transition from memorization to generalization in grokking.

In contrast, LAR depends on both weights and activations and is computable from matrix norms alone. It admits an explicit effective-dimension formula verified in grokking experiments, and tracks the generalization gap at 3B scale. Furthermore, effective rank only weakly distinguishes overfitting from generalization under Adam in 3B pre-training experiments (\cref{app:other-spectral-measures}), while unembedding LAR separates them cleanly under both Adam and Muon. 

\paragraph{Neural network parameterizations.}
A complementary line of work studies neural network \emph{parameterizations} as a principled
approach to transferring hyperparameters from small to large models.
The maximal update parameterization ($\mu$P) \citep{yang2021tensor,yang2024tensor} enables width transfer for learning rate and related   hyperparameters.
Subsequent work has extended this framework along new scaling axes:
\citet{dey2025completep} introduce CompleteP, which achieves hyperparameter transfer across
depth while also ensuring non-lazy learning in every layer, and
\citet{mlodozeniec2025completed} further extend the framework to cover per-module
hyperparameters, batch size, and token horizon.
\citet{everett2024scaling} revisited the alignment assumptions underlying the maximal update parameterization
and showed that relaxing them yields per-layer learning rate prescriptions that outperform $\mu$P
at scales up to 26.8B parameters.
LAR was introduced in \citet{everett2024scaling} as an empirical measure of parameter--activation
alignment as part of this analysis.
Our work repurposes LAR as a diagnostic for generalization: the same quantity that probes whether
alignment assumptions hold during training also tracks overfitting dynamics, independent of its uses in 
parameterization theory.

\paragraph{Grokking.}
Grokking, first identified by \citet{power2022grokking}, refers to the delayed transition from
memorization to generalization observed long after training accuracy has saturated.
Several lines of work have offered explanations. 
\citet{barak2022hidden} study hidden progress measures, i.e. scalar functions of the training 
state predictive of convergence time, and use sparse parity learning as a testbed in which 
grokking-style phase transitions arise and are consistent with their hidden-progress framework. 
\citet{nanda2023progress} mechanistically
reverse-engineer the generalizing solution for modular addition and define progress measures for
its emergence; \citet{liu2023omnigrok} attribute grokking to a mismatch between the geometry of
training and test loss landscapes; \citet{davies2022unifying} unify grokking and double descent as
instances of the same dynamic; and \citet{varma2023explaining} explain grokking as a competition
between a memorizing circuit and a more parameter-efficient generalizing circuit.
In the spectral-dynamics literature, \citet{yunis2024spectral} observe that the validation-loss
drop during grokking coincides with a sharp reduction in the effective rank of the weight
matrices.

Rather than propose a new theory, we use grokking as a testbed to study the behavior of unembedding LAR
during the memorization-to-generalization transition. We also find that LAR predicts the effective dimension
of the learned function, $k \approx n^{2(1-\LAR)}$, which we verify empirically
across 12 tasks.

\section{Log-Alignment Ratio}

\subsection{Definition} \label{sec:lar-definition}

Let $W \in \mathbb{R}^{m \times n}$ be a matrix, $X \in \mathbb{R}^{b \times n}$ be a batch of 
  input vectors, and $W X \in \mathbb{R}^{b \times m}$ be the batch of corresponding outputs. The 
  log-alignment ratio (LAR) of the matrix is defined by \cite{everett2024scaling} as:
\begin{equation} \label{eq:lar-orig}
\text{LAR} = \log_n \frac{\|WX\|_{\RMS} }{\|W\|_{\RMS} \cdot \|X\|_{\RMS}}
\end{equation}
where $\|\cdot \|_{\RMS}$ is the RMS-norm of the matrix.

    Let $W = \sum\limits_{i=1}^r u_i s_i v_i^\top$ be the SVD of $W$, where $r = \rank(W)$. Each     
  input vector $x \in \mathbb{R}^n$ can be expanded in the basis of right-singular 
  vectors as $x = \sum\limits_{i=1}^n x_{i} v_i$ with $x_{i} = v_i^\top x$, and $v_{r+1},    
  \ldots, v_n$ complete an orthonormal basis of $\mathbb{R}^n$ when $n > r$.

 We define the \emph{weight distribution} $p$ and the \emph{activation distribution} $q$ as
   the normalized squared singular values of $W$ and the normalized squared projections of $x$
   onto its singular directions, respectively:                                                
    \begin{equation}
    p_i = \frac{s_i^2}{\sum_{j=1}^r s_j^2}, \qquad
    q_i = \frac{\sum_{x \in B} (v_i^\top x)^2}{\sum_{x \in B}\|x\|^2}
    \label{eq:pq}
    \end{equation}                                                                        
                                                                                              
  where $\| \cdot \|$ denotes the $L_2$-norm of a vector. $p$ is a distribution over $r$ elements capturing how $W$ distributes its energy    
  across singular directions, and $q$ is a distribution over $n$ elements capturing how the   
  batch of activations distributes its variance along those directions and the null           
  space.\footnote{$q_i$ for $i = r+1, \ldots, n$ depends on the choice of basis vectors 
  completing the null space.}

A direct calculation gives                                                           
\begin{equation}                                                                          
\LAR = 1 + \frac{1}{2}\log_n\left(\sum_{i=1}^r p_i \, q_i\right).                          
\label{eq:lar-pq}
\end{equation}          

To see this, write 
\[
\|W\|_{\RMS}^2 = \frac{1}{nm}\sum\limits_{j=1}^r s_j^2 , \qquad \| X \|_{\RMS}^2 = \frac{1}{|B|\,n}\sum_{x \in B}\|x\|^2, \qquad \|Wx\|_{\RMS}^2= \frac{1}{|B|\,m}\sum_{x \in B}\sum\limits_{i=1}^r s_i^2\,x_i^2.
\]

The squared ratio becomes
\begin{align*}
  \frac{\|Wx\|_{\RMS}^2}{\|W\|_{\RMS}^2\;\|x\|_{\RMS}^2}
  &= \frac{\frac{1}{|B|\,m}\sum_{x \in B}\sum_i s_i^2\,x_i^2}
          {\frac{1}{nm}\sum_j s_j^2 \;\cdot\; \frac{1}{|B|\,n}\sum_{x \in B}\|x\|^2} \\
  &= n^2 \cdot \frac{\sum_i s_i^2 \sum_{x \in B} x_i^2}
                     {\sum_j s_j^2 \;\cdot\; \sum_{x \in B}\|x\|^2} \\
  &= n^2 \sum_{i=1}^r \frac{s_i^2}{\sum_j s_j^2}\cdot\frac{\sum_{x \in B} x_i^2}{\sum_{x \in B}\|x\|^2} \\
  &= n^2 \sum_{i=1}^r p_i\,q_i.
\end{align*}

Taking $\log_n$ of the square root gives the result in \cref{eq:lar-pq}.

Treating $p_i$ and $q_i$ as two $n$-vectors of equally-weighted data points (with $p_j = 0$ for $j > r$), we can write $\LAR$ in terms of their covariance, $\text{cov}(p, q) = \frac{1}{n} \sum\limits_{i=1}^r p_i q_i - \frac{1}{n^2}$,
\begin{equation}                                                                          
\LAR = 1 + \frac{1}{2}\log_n\left(n \, \text{cov}(p, q) + \frac{1}{n} \right).                      
\label{eq:lar-cov}
\end{equation}

Equivalently, in terms of the Pearson correlation coefficient, $\text{corr}(p, q) = \frac{\text{cov}(p, q)}{\sigma_p \, \sigma_q}$, where $\sigma_p$ and $\sigma_q$ are the standard deviations of $p$ and $q$ with $\sigma_p^2 = \frac{1}{n} \sum\limits_{i=1}^n p_i^2 - \frac{1}{n^2}$ and $\sigma_q^2 = \frac{1}{n} \sum\limits_{i=1}^n q_i^2 - \frac{1}{n^2}$, 
\begin{equation}                                                                          
\LAR = 1 + \frac{1}{2}\log_n\left(n \, \sigma_p \, \sigma_q \, \text{corr}(p, q) + \frac{1}{n} \right).                          
\label{eq:lar-corr}
\end{equation}

Notably, smaller values of $\sigma^2$ correspond to more diffuse distributions (see \cref{app:sigma-diffuse} for a derivation).

In practice, we compute LAR through \cref{eq:lar-orig} which has a small, negligible computational overhead of $O(|B|m + |B| n + mn)$. \cref{eq:lar-pq,eq:lar-cov,eq:lar-corr} provide alternative perspectives into the metric as overlap between two distributions. While LAR is well-defined for any matrix $W$, we focus on the unembedding matrix in \cref{sec:grokking,sec:large-scale}. Unless otherwise stated, "LAR" refers to the unembedding LAR in these sections. We discuss LAR for other matrices in \cref{app:lar-other-matrices}.

\subsection{Properties} \label{sec:lar-properties}
For any matrix $W$, LAR satisfies the following properties.
\begin{enumerate}
    \item $\LAR \in (-\infty, 1]$, with $\LAR = -\infty$ if activations lie entirely in the null space of~$W$, and $\LAR = 1$ if $W$ is a rank-1 matrix and activations lie entirely along its singular direction. In practice, backpropagation in neural networks encourages alignment between weights and activations, and we observe $\LAR \in (0.3, 0.8)$.
    \item Through \cref{eq:lar-cov}, $\LAR \approx \tfrac{1}{2}$ when $\operatorname{cov}(p, q) \approx 0$. Equivalently, \cref{eq:lar-corr} gives $\LAR \approx \tfrac{1}{2}$ whenever any of the following holds: the weight spectrum is uniform ($\sigma_p \approx 0$), the activations are isotropic ($\sigma_q \approx 0$), or $p$ and $q$ 
    are uncorrelated ($\operatorname{corr}(p, q) \approx 0$).
    Weights and activations of a neural network are uncorrelated at initialization and hence, $\LAR \approx \frac{1}{2}$.
    \item If $W$ has its singular values distributed uniformly across $k \leq r$ directions ($s_1 = \cdots = s_k > 0$, $s_i = 0$ for $i > k$) and all activation variance is contained in those $k$ directions, then
    \begin{equation}
      \LAR = 1 - \tfrac{1}{2}\log_n(k).
      \label{eq:lar-k}
    \end{equation}
    To see this, note $p_i = 1/k$ for $i \leq k$ and $q_i = 0$ for $i > k$ implies $\sum\limits_{i=1}^r p_i q_i = \tfrac{1}{k}\sum\limits_{i=1}^k q_i = 1/k$, and \Cref{eq:lar-k} follows. In this setup, we say that LAR captures the \textit{effective dimension} $k$ of the function learned by the neural network, with $k = n^{2(1-\LAR)}$. 
\end{enumerate}

\section{Grokking Experiments} \label{sec:grokking}

\begin{figure}[t]
  \centering
  \begin{subfigure}[t]{\columnwidth}
    \centering
    \includegraphics[width=\columnwidth]{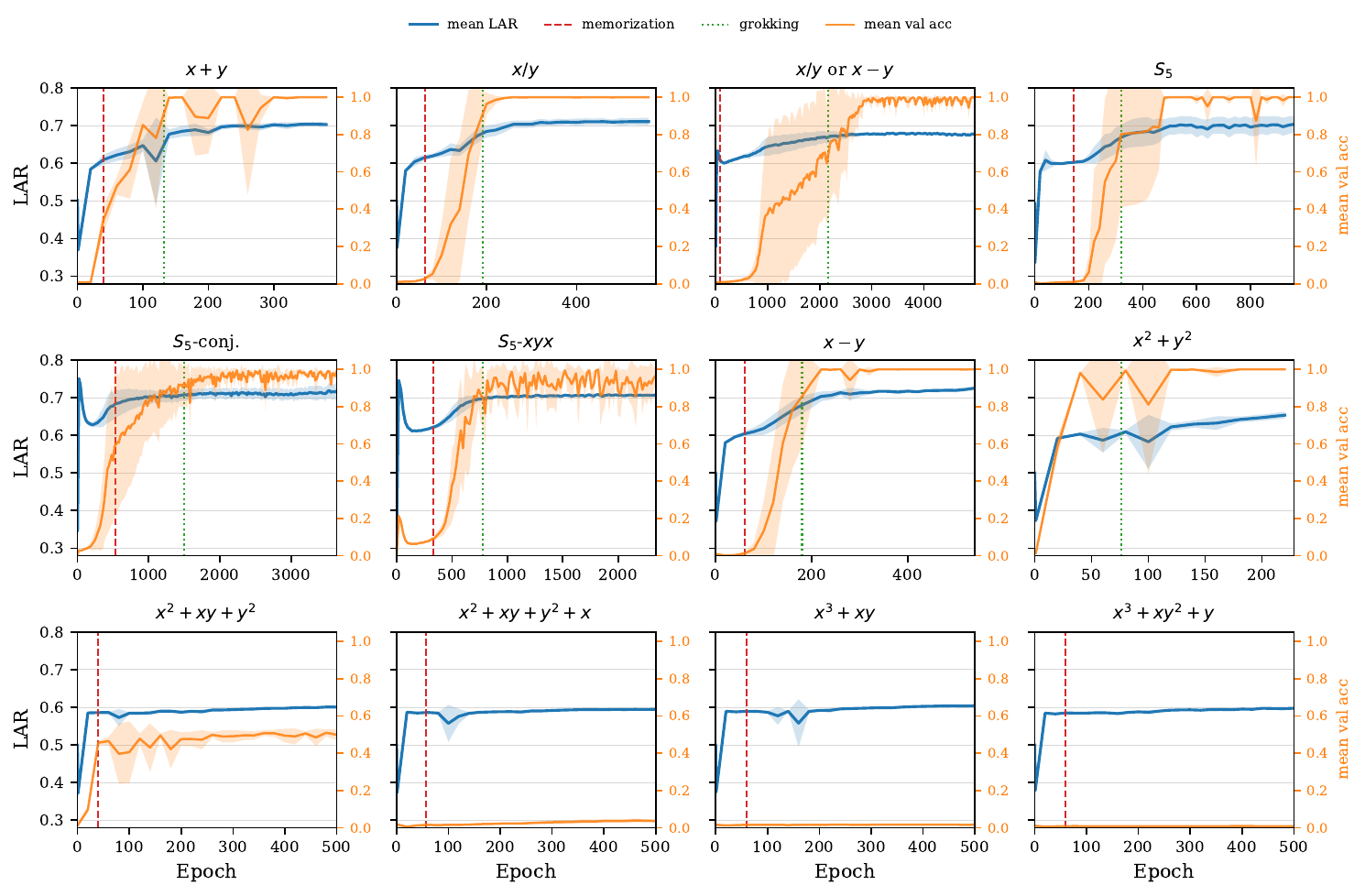}
    \caption{12 binary operation tasks.}
    \label{fig:grokking-12-tasks}
  \end{subfigure}
  \vspace{0.5em}
  \begin{subfigure}[t]{\columnwidth}
    \centering
    \includegraphics[width=\columnwidth]{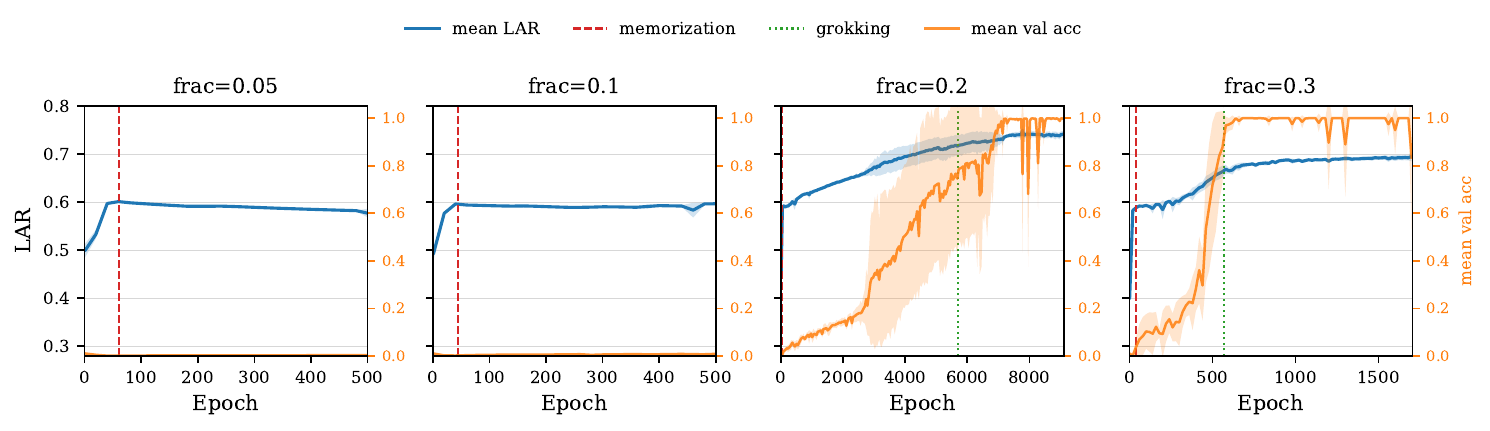}
    \caption{Modular addition with varying training dataset fraction.}
    \label{fig:grokking_dataset_fraction}
  \end{subfigure}
  \caption{Unembedding LAR (blue) and validation accuracy (orange) vs. epochs, with mean and ±1 std shaded bands across 5 seeds. Red and green dashed vertical lines mark   
  the mean memorization and grokking epochs. The absence of a green line indicates that these tasks do not grok.}
  \label{fig:grokking-lar-combined}
\end{figure}

\subsection{Setup}

Following \citet{power2022grokking}, we train 2-layer decoder-only Transformer models (with embedding dimension $128$, feed-forward dimension $512$, and $4$ attention heads) on $12$ binary operation tasks (cf. \cref{app:tasks}). We use AdamW \citep{loshchilov2019decoupled} with learning rate $ 10^{-3}$, weight decay $= 1.0$, and $\beta_2 = 0.98$, with batch size $512$, and $10$-step linear warmup. We train for up to $50,000$ epochs, terminating $2,000$ epochs after grokking if it occurs.

We conduct two sets of experiments. First, we train on all 12 tasks using a 50\% train     split. Second, we focus on modular addition ($x + y \pmod{97}$) and vary the training fraction across 5\%, 10\%, 20\%, and 30\% of the entire dataset to study how unembedding LAR behaves as a function of training set size. Each configuration is trained with 5 seeds. We compute validation accuracy every 20 epochs and check whether a model has memorized (training accuracy $> 99\%$, validation accuracy $< 50\%$) or grokked (training accuracy $> 99\%$, validation accuracy $> 99\%$). We also compute unembedding LAR at the same epochs using the entire training dataset rather than the current batch, as the dataset sizes are small in this setup.

\subsection{Experimental Results}

For our two sets of experiments, we plot the mean unembedding LAR and validation accuracy across seeds as functions of training epochs in \cref{fig:grokking-12-tasks} and \cref{fig:grokking_dataset_fraction}. We clip the range of x-axis to 3 times the grokking epoch if grokking occurs, else to $500$. 
In each case, we plot vertical dashed lines in red and green to represent the mean memorization and grokking epochs respectively. In 4 out of 12 cases in \cref{fig:grokking-12-tasks} and in 2 out of 4 cases in \cref{fig:grokking_dataset_fraction}, the green line is missing as these models do not grok.

  In all cases, we make the following observations. 
\begin{enumerate}
  \item  LAR starts at $\sim 0.5$ at initialization, as expected. It then decreases briefly before increasing steadily  and stabilizing in the range $0.6$--$0.72$ at the end of training. 
 \footnote{The initial drop is shown more clearly in \cref{fig:early_lar_dip} in \cref{app:lar-batch}, where we show that it stems from using a large batch size.} LAR at memorization is lower than at grokking, which is lower than at the final checkpoint. 
 \item Grokking models achieve higher values ($\LAR > 0.65$) than non-grokking models ($\LAR \leq 0.65$).    
 \end{enumerate}

\begin{figure}[t]
  \centering
  \begin{subfigure}[t]{0.57\textwidth}
    \centering
    \includegraphics[width=\textwidth]{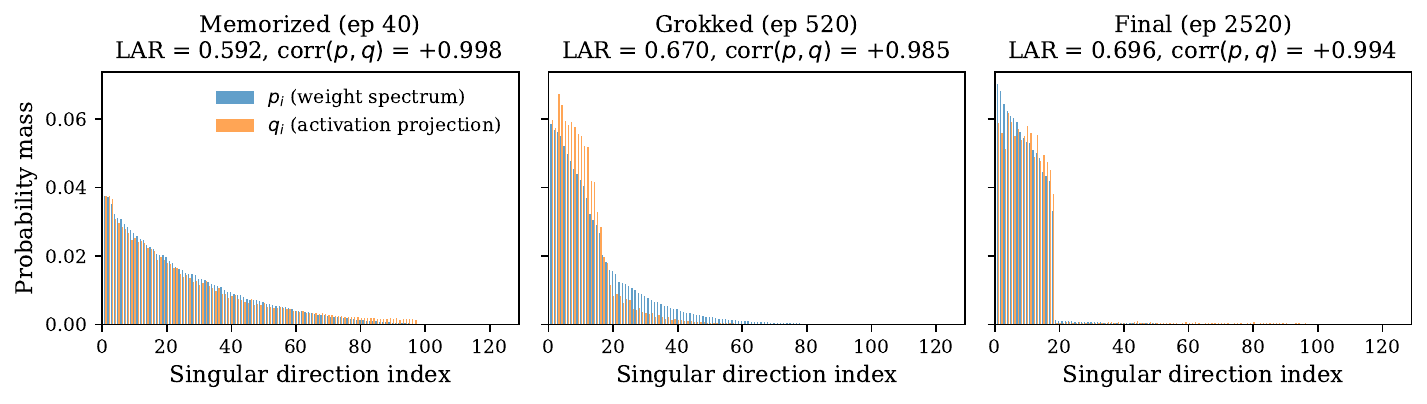}
    \caption{Modular addition, $30\%$ training data}
    \label{fig:grokking_add_30p}
  \end{subfigure}
  \hfill
  \begin{subfigure}[t]{0.4\textwidth}
    \centering
    \includegraphics[width=\textwidth]{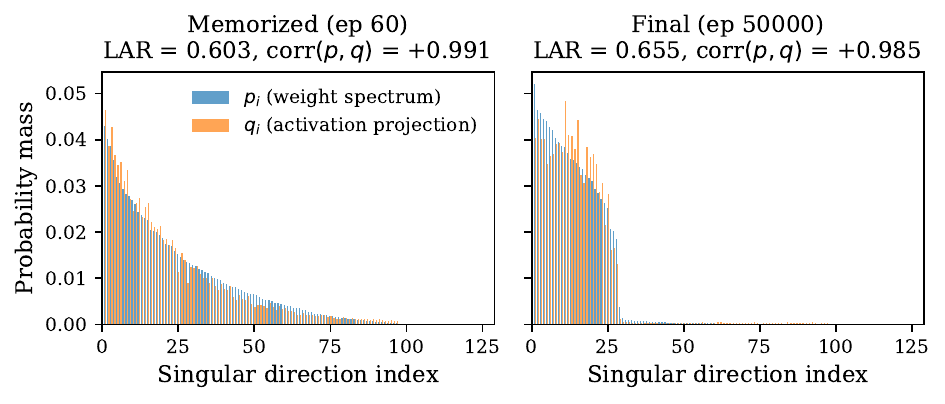}
    \caption{Modular addition, $5\%$ training data}
    \label{fig:grokking_add_5p}
  \end{subfigure}
  \vspace{0.5em}
  \begin{subfigure}[t]{0.57\textwidth}
    \centering
    \includegraphics[width=\textwidth]{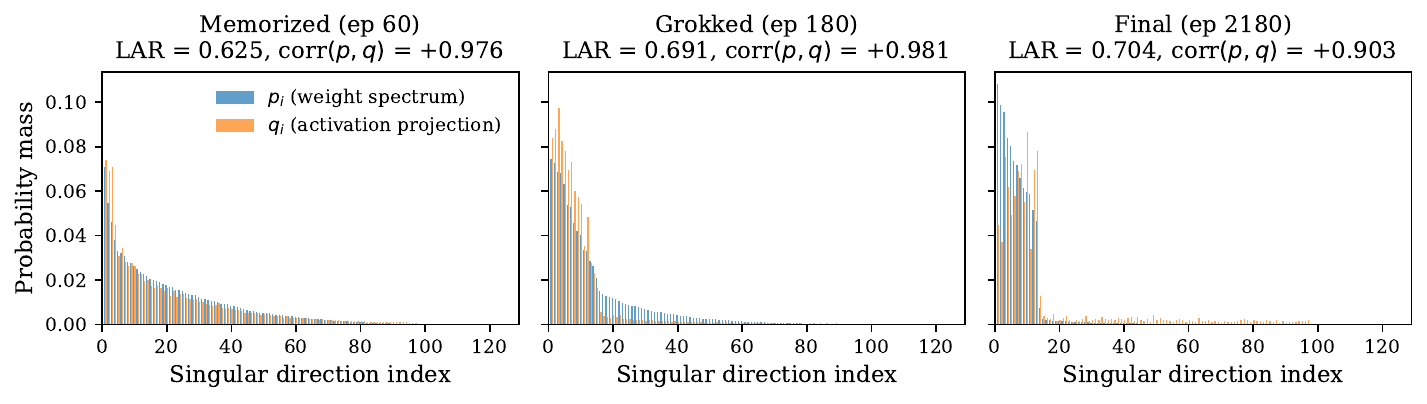}
    \caption{Modular Division}
    \label{fig:grokking_div}
  \end{subfigure}
  \hfill
  \begin{subfigure}[t]{0.4\textwidth}
    \centering
    \includegraphics[width=\textwidth]{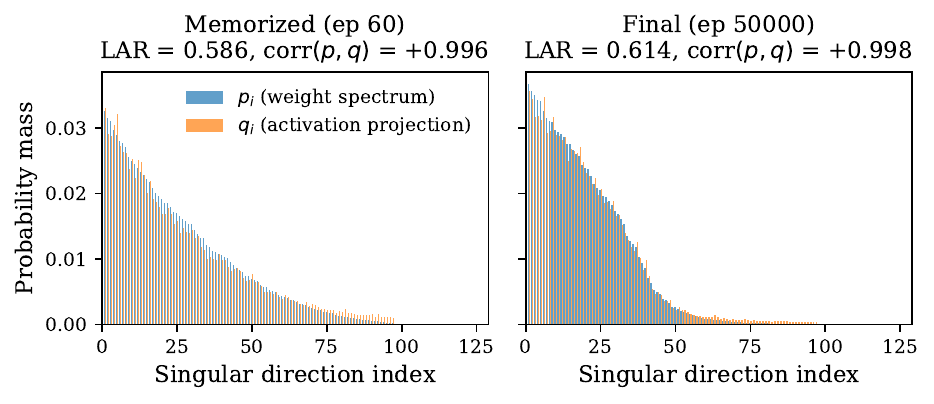}
    \caption{$x^3 + xy^2 + y \pmod{97}$}
    \label{fig:grokking_x3}
  \end{subfigure}
  \caption{Weight distribution $p$ and activation distribution $q$ at different points during training. Memorization, grokking, and final points are shown for tasks that grok in (a) and (c). Only memorization and final points are shown for tasks that do not grok in (b) and (d). Subplot titles include epoch, $\LAR$, and $\corr(p, q)$ values.}
  \label{fig:grokking-pq-combined}
\end{figure}

\subsection{Analysis}

Motivated by the interpretation of LAR as the overlap between the weight distribution $p$ and activation distribution $q$, we plot $p$ and $q$ at memorization, grokking, and final checkpoints for four tasks in \cref{fig:grokking-pq-combined}. In each case, $p$ and $q$ are diffuse at memorization, concentrated at grokking, and most concentrated at the final checkpoint. Equivalently, $\sigma_p$ and $\sigma_q$ are smallest at memorization, larger at grokking, and largest at the final checkpoint.
 Since $\operatorname{corr}(p, q)$ remains near $1$ throughout, \cref{eq:lar-corr} then explains the evolution of LAR during training: it is smaller at memorization due to more diffuse $p$ and $q$, and larger at grokking and final checkpoints due to more concentrated distributions.

\begin{figure}[t]
  \centering
  \begin{subfigure}[t]{0.48\textwidth}
    \centering
    \includegraphics[width=\textwidth]{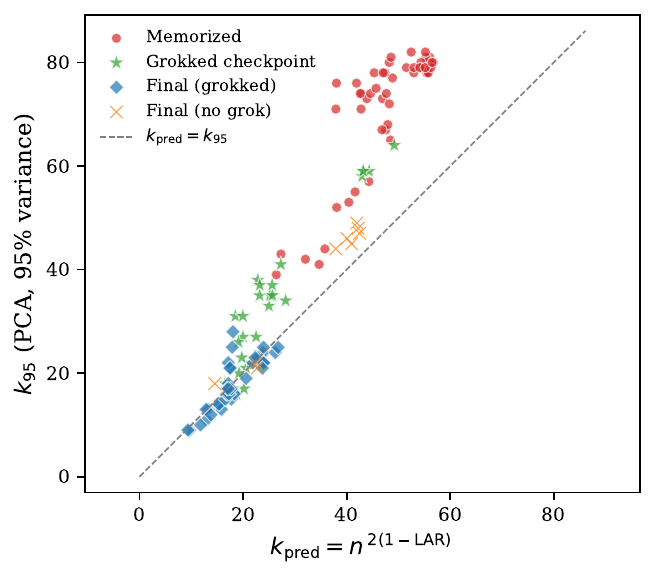}
    \caption{12 binary operation tasks.}
    \label{fig:kpred_v_k95_tasks}
  \end{subfigure}
  \hfill
  \begin{subfigure}[t]{0.48\textwidth}
    \centering
    \includegraphics[width=\textwidth]{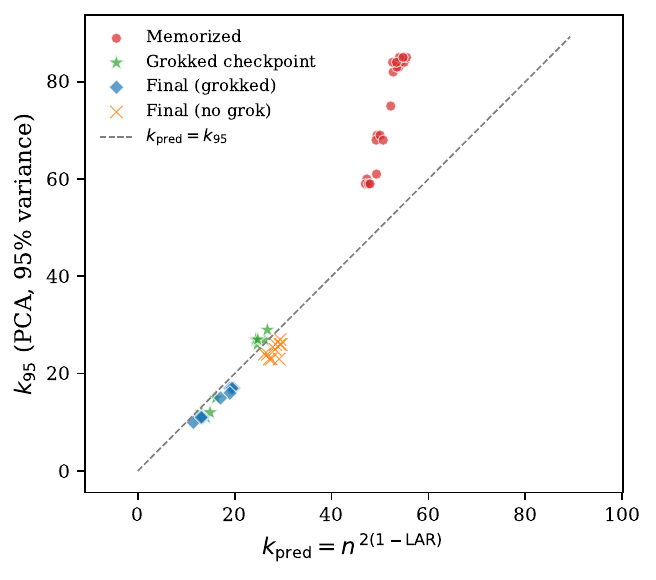}
    \caption{Modular addition, varying train fraction.}
    \label{fig:kpred_v_k95_frac}
  \end{subfigure}
  \caption{$k_{95}$ (the number of principal components explaining $95\%$ of unembedding input activation variance) vs.\ LAR-predicted      
  dimensionality $k_p = n^{2(1-\LAR)}$. Each point corresponds to a memorization, grokking or final checkpoint.} 
  \label{fig:kpred_v_k95_split}
\end{figure}

We hypothesize that optimization forces $p$ and $q$ to concentrate on the same small subset of directions, across all tasks and seeds in our experiments. We test this hypothesis by performing PCA on the input activations of the unembedding matrix and computing the number of principal components, $k_{95}$, that explain $95\%$ of the variance. $k_{95}$ is the number of directions along which $q$ is mostly concentrated. If $p$ is concentrated along the same $k_{95}$ directions, property 3 from \cref{sec:lar-properties} can provide a good approximation.
\footnote{Strictly, property 3 requires $p$ to be \textit{uniform} across $k$ directions, which does not hold exactly in our experiments.     
 }
In this case, $\LAR \approx 1 - \frac{1}{2} \log_n (k_{95})$, or equivalently, $k_{\pred} := n^{2(1-\LAR)}$ approximates $k_{95}$.

We plot $k_{95}$ and $k_{\pred}$ for grokking, memorization and final checkpoints of each individual experiment in \cref{fig:kpred_v_k95_split}. We observe that the approximation holds well for final and (to some extent) grokking checkpoints, but $k_{95} > k_\pred$ invariably at memorization. The root-mean-squared-error values, $\|k_{95} - k_\pred \|$, given in \cref{tab:kpred-rmse} support this observation. Larger values of $k_{95}$ at memorization reflect more diffuse $q$, while its smaller values and better approximationss to $k_\pred$ at grokking and final checkpoints indicate $p$ and $q$ becoming concentrated on the same small subset of directions. 

Lastly, we note that $k_{95} \approx k_\pred$ takes larger values for final checkpoints of non-grokking models as compared to grokking models. Thus, a grokking model learns a more compressed representation than its non-grokking counterpart, which is reflected in grokking models achieving higher values of LAR.

  \begin{table}[t]
    \caption{RMSE between $k_p$ and $k_{95}$ at memorized, grokked, and final checkpoints. Values are RMSE $\pm$ half the 95\% percentile bootstrap CI (10{,}000 resamples).}
    \label{tab:kpred-rmse}                                                                         
    \centering                                                                                                                                                      
    \small                                                                                                                                                          
    \begin{tabular}{@{}l cccc@{}}                                                                                                                                   
    \toprule      
    Experimental Setup & Memorized & Grokked & Final (grokked) & Final (no grok) \\
    \midrule                                                                                                                                                        
    $12$ binary operations
      & $24.90 \pm 1.67$                                                                                                                                            
      & $8.62 \pm 1.70$                                                                                                                                             
      & $2.70 \pm 1.03$
      & $4.80 \pm 1.10$ \\                                                                                                                                          
    Modular addition (train fraction sweep)
      & $24.33 \pm 3.85$                                                                                                                                            
      & $2.19 \pm 0.92$
      & $2.08 \pm 0.34$                                                                                                                                             
      & $3.21 \pm 0.55$ \\                                                                                                                                          
    \bottomrule
    \end{tabular}                                                                                                                                                   
  \end{table}

\section{Large-Scale Experiments} \label{sec:large-scale}
\subsection{Experimental Setup}

We train 3B-parameter models with embedding dimension $4096$ and Gemma-2 decoder blocks \citep{gemma2024} using two optimizers, Muon \citep{jordan2024muon} and Adam \citep{kingma2015adam}, across five dataset sizes: 200M, 400M, 800M, 1.6B, and 100B tokens. Each model is trained for 2B tokens, so the number of epochs varies with dataset size. The smaller datasets are random subsets of the largest, ensuring similar distributions; in each case, 10\% is held out for validation. The training batch size is 24 sequences of length 8{,}192 (196{,}608 tokens), and we evaluate 
  validation loss every 10 steps on a batch of 2.6M tokens. We sweep over learning rates on the 100B dataset to find the optimal rate for each optimizer, then use it for all other dataset sizes. Further details are provided in \cref{app:exp-details-large-scale}.

\subsection{Results}
\begin{figure}[t]
    \centering                                          
    \includegraphics[width=\textwidth]{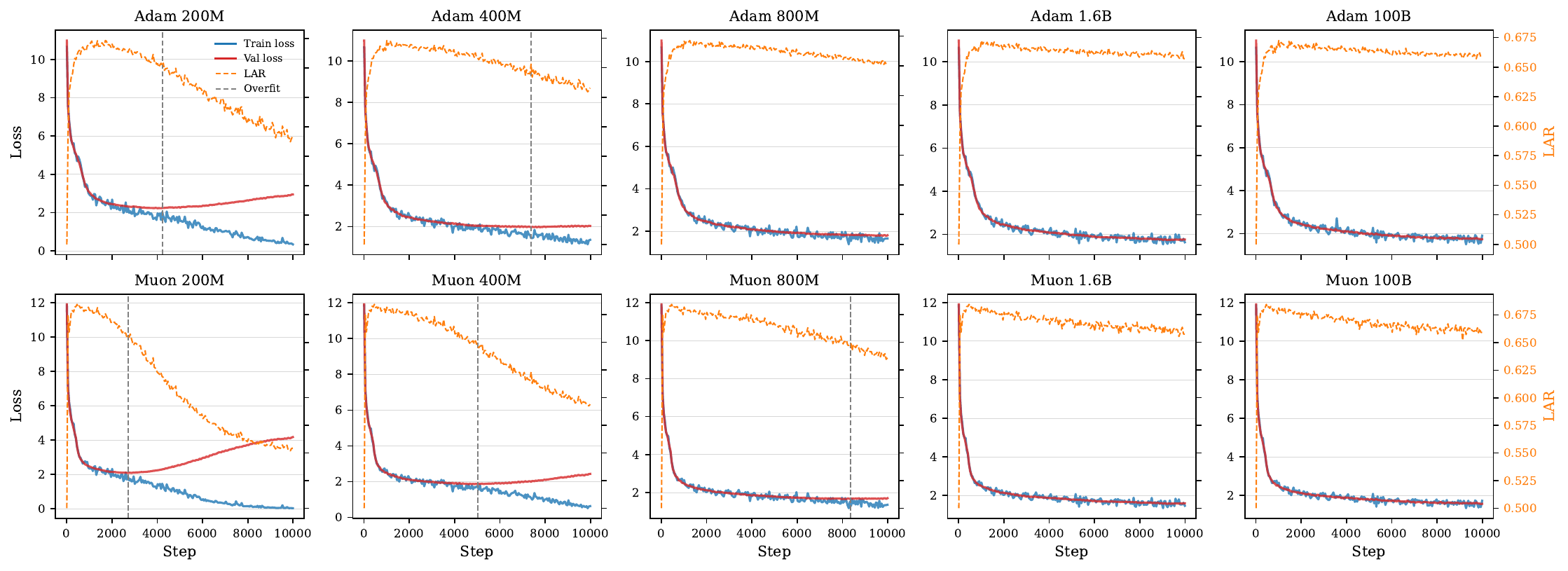}
    \caption{Training loss, validation loss, and unembedding log-alignment ratio (LAR) as functions of training steps. Vertical dashed lines represent the training step with minimum validation loss.}                                                              
    \label{fig:train_val_lar}
  \end{figure}

\begin{figure}[t]
    \centering                                          
    \includegraphics[width=\textwidth]{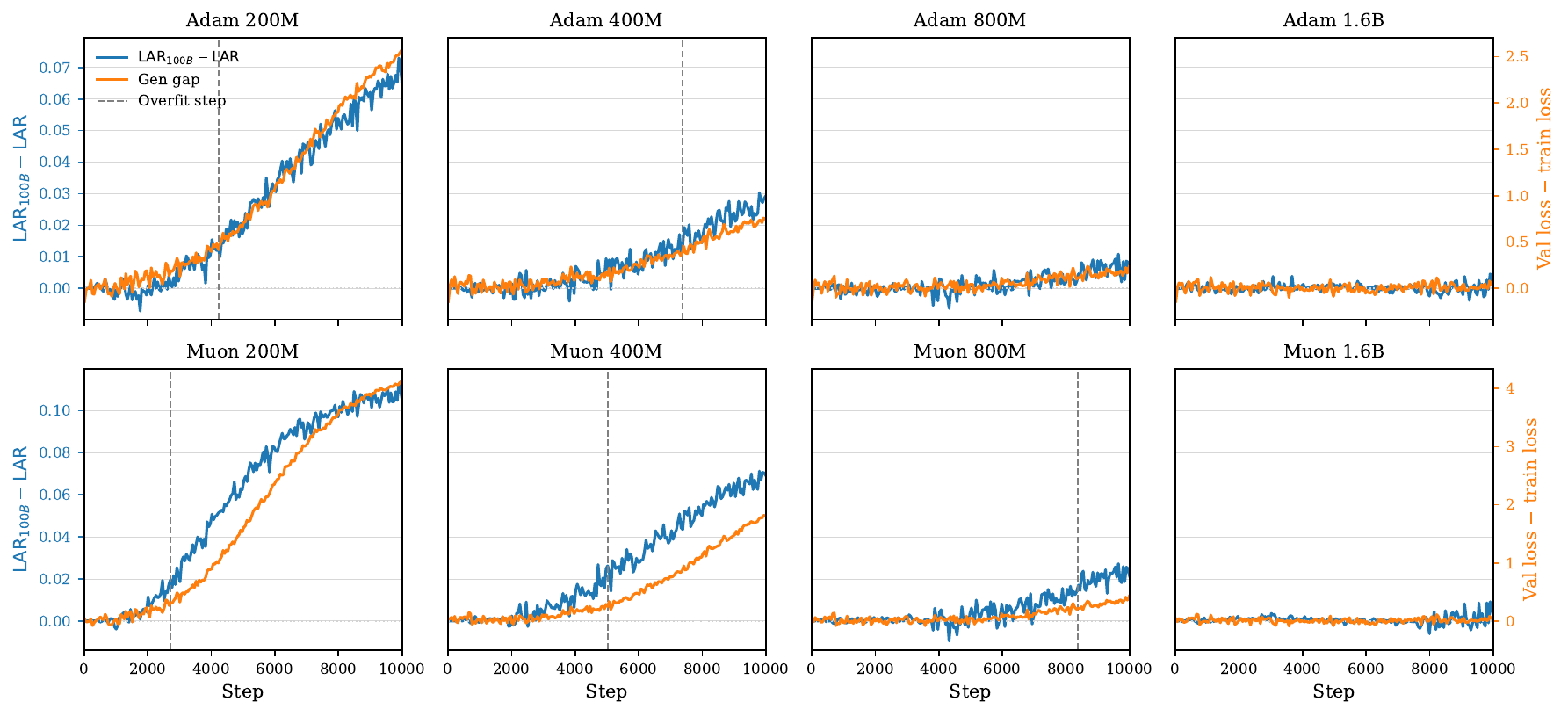}
    \caption{LAR differential, defined as the difference between the baseline LAR from the $100\text{B}$ experiment and the LAR for a given dataset size, and generalization gap — the difference between validation loss and smoothed training loss as functions of training steps.}
    \label{fig:gengap_lar}
  \end{figure}

For each of the 10 models in our setup, \cref{fig:train_val_lar} shows training loss, validation loss, and unembedding LAR, with the vertical dashed lines marking the step of minimum validation loss. 
\footnote{We evaluated several criteria for detecting the onset of overfitting, aiming for one
       that triggered as early as possible without showing any false positives on the 100B baseline. 
     The best rule we found was simple: flag the first step at which  
    the mean validation loss over a 5-step window exceeds the running minimum by at     
    least 2\%. In practice, this rule identifies essentially the same step as the minimum
    of the validation loss curve itself, which we attribute to our large validation    
   batch (2.6M tokens) producing a smooth loss trace with little step-to-step noise.} 
2 out of 5 models overfit with Adam, and 3 out of 5 models overfit with Muon. 

The case of the largest dataset size, 100B, establishes a baseline behavior of LAR for non-overfitting models. 
In this case, LAR starts at $\sim$0.5 at initialization, as expected, then rises to a peak of  $\sim$0.676 with Adam and $\sim$0.688 with Muon before slowly declining throughout training.
\footnote{We suspect that the initial rise in LAR is due to the use of large optimization batch size. We leave a verification of this hypothesis to future work.}
In contrast, overfitting models exhibit larger decline in LAR towards the end of training.

\begin{table}[t]
    \centering
    \caption{Rate of decline in unembedding LAR in 1000-step windows after the peak step, measured as negative slope per 1000 steps. 
    Higher values indicate steeper decline and are visualized as darker cell color.
    \textbf{Bold} values indicate up to three windows ending at the detected overfitting step.
    100B and 1.6B runs show slowdown in decline towards the end of training, while overfitting
    runs show sharp decline before the overfitting step.}
    \label{tab:lar-decline-1k}
    \resizebox{\textwidth}{!}{%
    \begin{tabular}{lrrrrr@{\hspace{1.5em}}rrrrr}
    \toprule
    & \multicolumn{5}{c}{\textbf{Adam}} & \multicolumn{5}{c}{\textbf{Muon}} \\
    \cmidrule(lr){2-6} \cmidrule(lr){7-11}
    \textbf{Window} & \textbf{100B} & \textbf{1.6B} & \textbf{800M} & \textbf{400M} & \textbf{200M}
                    & \textbf{100B} & \textbf{1.6B} & \textbf{800M} & \textbf{400M} & \textbf{200M} \\
    \midrule
    1001 $\to$ 2000    & \cellcolor{orange!32!white}0.0014 & \cellcolor{orange!32!white}0.0014 &
  \cellcolor{orange!36!white}0.0020 & \cellcolor{orange!27!white}0.0009 & \cellcolor{orange!31!white}0.0013 &
  \cellcolor{orange!44!white}0.0036 & \cellcolor{orange!48!white}0.0050 & \cellcolor{orange!46!white}0.0043 &
  \cellcolor{orange!41!white}0.0029 & \cellcolor{orange!57!white}\textbf{0.0113} \\
    2001 $\to$ 3000    & \cellcolor{orange!33!white}0.0015 & \cellcolor{orange!38!white}0.0022 &
  \cellcolor{orange!35!white}0.0018 & \cellcolor{orange!34!white}0.0017 & \cellcolor{orange!48!white}\textbf{0.0051} &
   \cellcolor{orange!43!white}0.0033 & \cellcolor{orange!46!white}0.0043 & \cellcolor{orange!40!white}0.0026 &
  \cellcolor{orange!52!white}0.0072 & \cellcolor{orange!63!white}\textbf{0.0183} \\
    3001 $\to$ 4000    & \cellcolor{orange!29!white}0.0011 & \cellcolor{orange!24!white}0.0007 &
  \cellcolor{orange!25!white}0.0008 & \cellcolor{orange!43!white}0.0035 & \cellcolor{orange!53!white}\textbf{0.0076} &
   \cellcolor{orange!41!white}0.0030 & \cellcolor{orange!34!white}0.0017 & \cellcolor{orange!37!white}0.0021 &
  \cellcolor{orange!55!white}\textbf{0.0094} & \cellcolor{orange!68!white}0.0258 \\
    4001 $\to$ 5000    & \cellcolor{orange!24!white}0.0007 & \cellcolor{orange!29!white}0.0011 &
  \cellcolor{orange!30!white}0.0012 & \cellcolor{orange!42!white}0.0031 & \cellcolor{orange!56!white}\textbf{0.0103} &
   \cellcolor{orange!41!white}0.0030 & \cellcolor{orange!38!white}0.0022 & \cellcolor{orange!45!white}0.0041 &
  \cellcolor{orange!59!white}\textbf{0.0123} & \cellcolor{orange!66!white}0.0221 \\
    5001 $\to$ 6000    & \cellcolor{orange!38!white}0.0022 & \cellcolor{orange!34!white}0.0017 &
  \cellcolor{orange!38!white}0.0022 & \cellcolor{orange!44!white}\textbf{0.0038} & \cellcolor{orange!58!white}0.0116 &
   \cellcolor{orange!43!white}0.0033 & \cellcolor{orange!44!white}0.0037 & \cellcolor{orange!50!white}0.0060 &
  \cellcolor{orange!59!white}\textbf{0.0124} & \cellcolor{orange!63!white}0.0179 \\
    6001 $\to$ 7000    & \cellcolor{orange!28!white}0.0010 & \cellcolor{orange!25!white}0.0008 &
  \cellcolor{orange!40!white}0.0026 & \cellcolor{orange!50!white}\textbf{0.0059} & \cellcolor{orange!57!white}0.0105 &
   \cellcolor{orange!38!white}0.0023 & \cellcolor{orange!38!white}0.0022 & \cellcolor{orange!49!white}\textbf{0.0058}
  & \cellcolor{orange!60!white}0.0141 & \cellcolor{orange!59!white}0.0129 \\
    7001 $\to$ 8000    & \cellcolor{orange!8!white}0.0002 & \cellcolor{orange!0!white}$-$0.0001 &
  \cellcolor{orange!32!white}0.0014 & \cellcolor{orange!48!white}\textbf{0.0050} & \cellcolor{orange!56!white}0.0096 &
   \cellcolor{orange!22!white}0.0006 & \cellcolor{orange!0!white}$-$0.0005 &
  \cellcolor{orange!49!white}\textbf{0.0055} & \cellcolor{orange!57!white}0.0111 & \cellcolor{orange!53!white}0.0080
  \\
    8001 $\to$ 9000    & \cellcolor{orange!13!white}0.0003 & \cellcolor{orange!20!white}0.0005 &
  \cellcolor{orange!39!white}0.0025 & \cellcolor{orange!48!white}0.0050 & \cellcolor{orange!53!white}0.0079 &
  \cellcolor{orange!0!white}$-$0.0003 & \cellcolor{orange!29!white}0.0011 & \cellcolor{orange!50!white}\textbf{0.0061}
   & \cellcolor{orange!54!white}0.0083 & \cellcolor{orange!47!white}0.0048 \\
    9001 $\to$ end     & \cellcolor{orange!13!white}0.0003 & \cellcolor{orange!22!white}0.0006 &
  \cellcolor{orange!34!white}0.0016 & \cellcolor{orange!43!white}0.0033 & \cellcolor{orange!51!white}0.0065 &
  \cellcolor{orange!25!white}0.0008 & \cellcolor{orange!32!white}0.0014 & \cellcolor{orange!48!white}0.0052 &
  \cellcolor{orange!52!white}0.0072 & \cellcolor{orange!41!white}0.0028 \\
    \midrule
    Peak step          & 887 & 1222 & 1104 & 1222 & 887 & 546 & 416 & 416 & 416 & 619 \\
    \bottomrule
    \end{tabular}%
    }
  \end{table}

In \cref{fig:gengap_lar}, we plot the LAR differential, defined as the difference between the baseline LAR (from the non-overfitting $100\text{B}$ experiment) and the LAR for a given dataset size, alongside the generalization gap — the difference between validation loss and smoothed training loss.
\footnote{We smooth training loss by averaging over a centered 14-step sliding window, approximately matching the 2.6M tokens used to compute validation loss.} 
We note that this differential has a similar shape to the generalization gap during training. 
The LAR differential therefore serves as a signal of overfitting that requires no validation data.

In \cref{tab:lar-decline-1k}, we further give the rate of decline in LAR over periods of $1000$ steps in training. 
 Non-overfitting models exhibit a steady rate of decline after the peak, followed by a stabilization phase towards the end of training.  
 In contrast, overfitting models show steeper declines as the overfitting step approaches.
 Notably, Adam 800M maintains elevated decline rates throughout training without formally overfitting in our budget of 2B tokens, suggesting that it is approaching the overfitting step. 
 The rate of change in slope of LAR, i.e., its acceleration, can therefore be a useful metric to check for overfitting ---  a possibility we leave open for future investigation.

\subsection{Analysis}

To understand why unembedding LAR tracks generalization, we compute $p$ and $q$ explicitly every $1000$ training steps during training. Across all experiments, the top singular value is 3 times as large as the next value, so $p_1 \approx 9 p_2$. (See \cref{app:pq-evolve} for detailed plots of $p$ and $q$ and related analysis.)

$\LAR$ admits a natural decomposition, $\LAR = L_k + \Delta_k$, into the \textit{cumulative} contribution from top $k$ singular directions, $L_k$, and the residual, $\Delta_k$: 
\[
L_k = 1 + \frac{1}{2} \log_n \left(\sum\limits_{i=1}^k p_i q_i \right) \quad ; \quad \Delta_k = - \frac{1}{2} \log_n \left( \frac{\sum\limits_{i=1}^k p_i q_i}{\sum\limits_{i=1}^r p_i q_i} \right)
\]

\begin{figure}[t]             
    \centering                                                                                                                        
    \begin{subfigure}[t]{0.48\columnwidth}                                                                                            
      \centering                                                                                                                      
      \includegraphics[width=\textwidth]{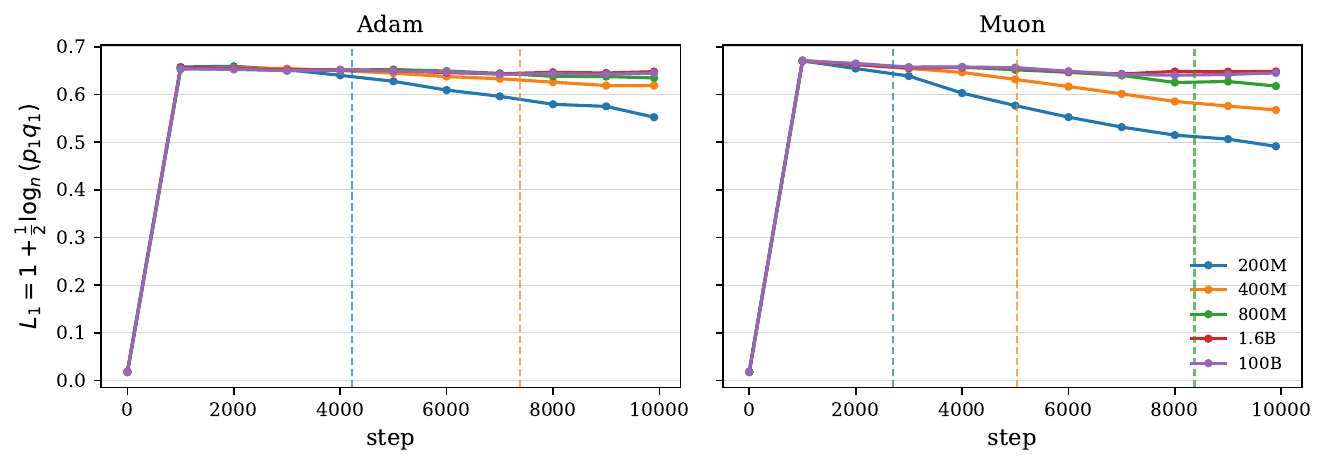}                                                            
      \caption{$L_1$}
      \label{fig:L1_1x2_overfit}
    \end{subfigure}                                                                                                                   
    \hfill        
    \begin{subfigure}[t]{0.48\columnwidth}
      \centering                                                                                                                      
      \includegraphics[width=\textwidth]{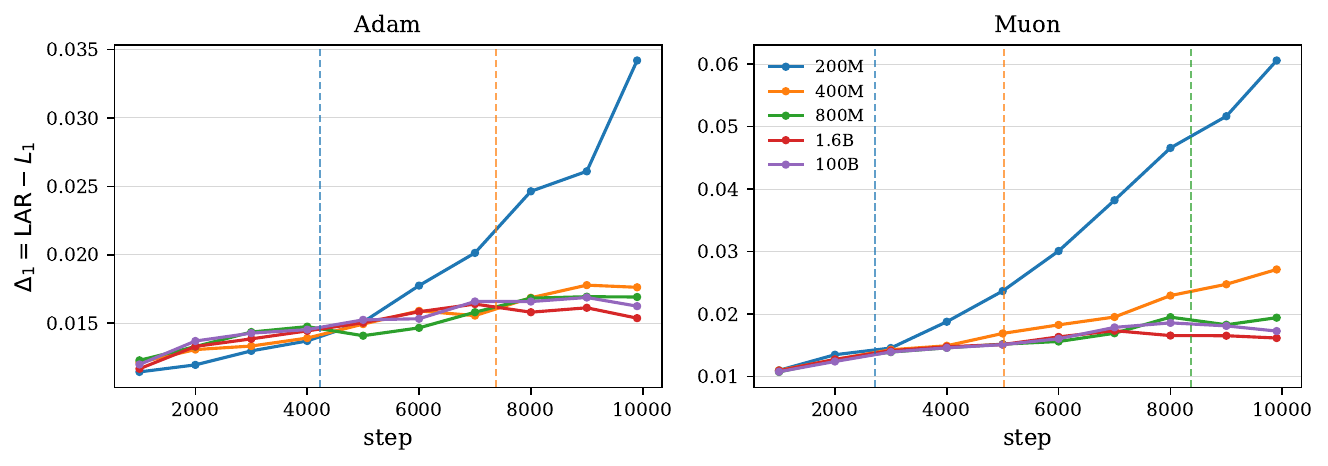}
      \caption{$\Delta_1$}  
      \label{fig:Delta_1x2_overfit_frompeak}
    \end{subfigure}                                                                                                                   
    \caption{Contribution of the top singular direction, $L_1$, to $\LAR$ (a), and the residual $\Delta_1 = \LAR - L_1$ (b), as functions of training steps. Different colors indicate dataset sizes; vertical dashed lines mark the onset of overfitting for the corresponding run. Step 0 is omitted in (b) for clarity.}
    \label{fig:l1_and_delta1}                                                                                             
  \end{figure}

\begin{figure}[t]                      
    \centering                                                                                                
    \includegraphics[width=\textwidth]{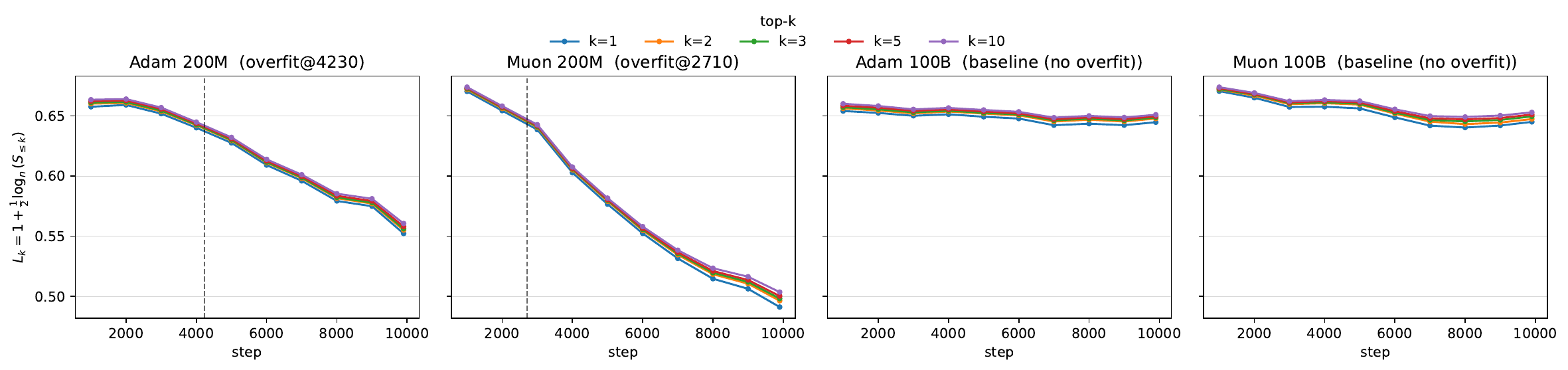}                            
    \caption{Contribution of top $k$ singular directions, $L_k$, to the unembedding LAR for $k=1, 2, 3, 5, 10$ in 200M and 100B experiments with Adam and Muon. Step 0 is omitted for clarity in all cases.}                                                 
    \label{fig:Lk_family_from_step1000}                                                                                         
  \end{figure}

We plot the evolution of $L_1$ and $\Delta_1$ in \cref{fig:l1_and_delta1}. $L_1$ diverges from the non-overfitting baseline prior to the overfitting step, while $\Delta_1$ stays approximately constant through the overfitting step. The pre-overfit signature of overfitting is thus contained mostly in $L_1$. This is further confirmed in \cref{fig:Lk_family_from_step1000}, where we plot $L_k$ for $k=1, 2, 3, 5, 10$ across 200M and 100B experiments. The curves for $k>1$ differ from $L_1$ only by a near-constant offset, following the same trend otherwise. 

Our observations are consistent with \citet{yang2023spectral}, who note that in the presence of feature learning, gradient descent induces alignment between a layer's input activations and the top singular 
directions of its weight matrix. The decline in $L_1$ before overfitting observed here reflects a weakening 
of this alignment.

\begin{figure}[t]             
    \centering                                                                                                                        
    \begin{subfigure}[t]{0.48\columnwidth}                                                                                            
      \centering                                                                                                                      
      \includegraphics[width=\textwidth]{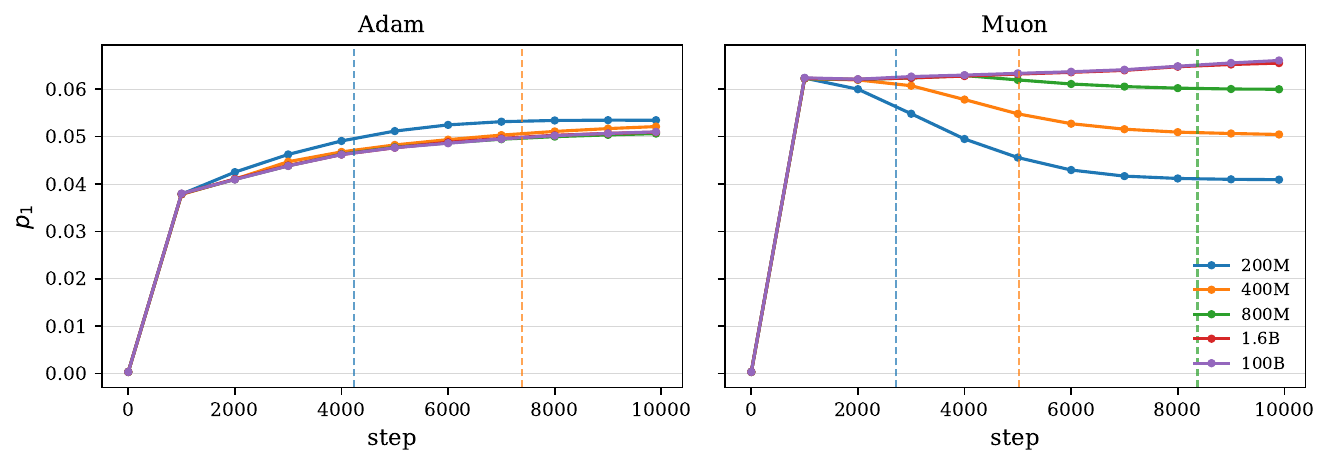}                                                            
      \caption{$p_1$}
      \label{fig:p1_1x2_overfit}
    \end{subfigure}                                                                                                                   
    \hfill        
    \begin{subfigure}[t]{0.48\columnwidth}
      \centering                                                                                                                      
      \includegraphics[width=\textwidth]{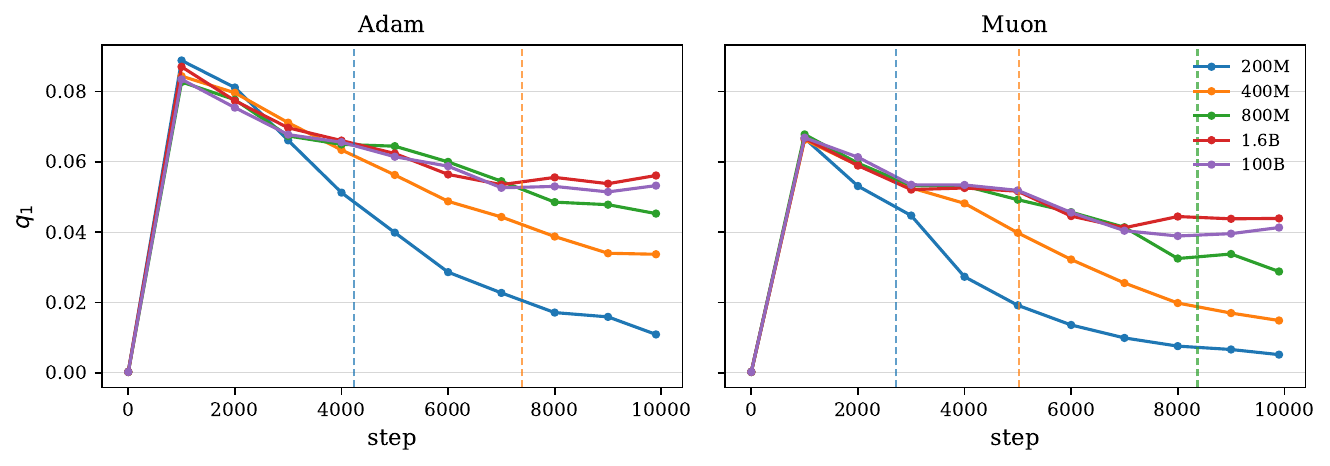}
      \caption{$q_1$}  
      \label{fig:q1_1x2_overfit}
    \end{subfigure}                                                                                                                   
    \caption{Weight-spectrum component $p_1$ (a) and activation-spectrum component $q_1$ (b) along the top singular direction, as functions of training steps. Different colors indicate dataset sizes; vertical dashed lines mark the onset of overfitting for the corresponding run.}
    \label{fig:p1_and_q1}                                                                                             
  \end{figure}

To further understand this weakening, we plot $p_1$ and $q_1$ separately in \cref{fig:p1_and_q1}. We note that with Adam, the signature of overfitting is  entirely in $q_1$, with $p_1$ exhibiting growth across all dataset sizes. With Muon, it is contained in both $p_1$ and $q_1$. Notably, the unembedding layer is optimized with Adam in both cases. We leave a mechanistic explanation of this asymmetry for future work.

\paragraph{Interpretation in terms of spreads in $p$ and $q$.} In grokking experiments of \cref{sec:grokking}, we observed that memorization corresponds to more diffuse distributions $p$ and $q$, while generalization corresponds to more concentrated distributions. We make a similar observation here: when $p_1$ and $q_1$ are much larger than their next counterparts, they dictate the spread in distributions as $\sigma_p \approx \frac{p_1}{\sqrt{n}}$ and $\sigma_q \approx \frac{q_1}{\sqrt{n}}$. With Adam, overfitting thus correlates with more diffuse activations and with Muon, it correlates with more diffuse activations and weight spectrum. We discuss the behavior of $\sigma_p$ and $\sigma_q$ in more detail in \cref{app:pq-evolve}.

\section{Discussion}

We have studied the unembedding log-alignment ratio (LAR) in two settings. On small algorithmic tasks, the final value of LAR tracks the effective dimensionality $k_{95}$ used by the model, which can be interpreted as the dimensionality of the generalizing circuit when the model groks. 
On 3B-parameter language models, LAR decreases slowly and eventually stabilizes when the model generalizes, but declines sharply once overfitting sets in.
LAR is computed entirely from training-time quantities with negligible computational overhead, and hence, does not depend on the evaluation data, making it a practically useful signal in situations where validation is expensive: for example, large-scale pretraining with limited evaluation budgets.

The decomposition $\LAR = 1 + \tfrac{1}{2}\log_n(n\,\sigma_p\,\sigma_q\,\operatorname{corr}(p,q) + 1/n)$ explains \emph{why} LAR declines during training: either because the weight or activation spectra become more diffuse, or because the two stop being concentrated on the same directions. Empirically, we observe that $p$ and $q$ remain highly correlated in almost all cases, and the behavior of $\LAR$ is explained by $\sigma_p \sigma_q$.
In pre-training experiments, the top singular value of the unembedding is much larger than the rest, so $\sigma_p\sigma_q$ further reduces to the top-direction overlap $p_1 q_1$, and the pre-overfit signature lives almost entirely in the top-component contribution $L_1 = 1 + \tfrac{1}{2}\log_n(p_1 q_1)$.
The behavior depends on the optimizer: with Adam, overfitting is reflected almost entirely in $\sigma_q$, while with Muon both distributions $p$ and $q$ spread out. Understanding this asymmetry, which is likely tied to how each optimizer propagates spectral structure through the rest of the network, is a natural direction for follow-up work.

\paragraph{Limitations.}
First, we focus only on the unembedding layer. While Appendix~\ref{app:lar-other-matrices} shows that some other matrices exhibit related properties, a complete understanding of LAR for each matrix type and the information it provides about the training dynamics is a promising direction for future work.
Second, while we vary the dataset size and observe LAR behavior in both the overfitting ($D \ll N$) and non-overfitting ($D \gg N$) regimes, our 100B baseline trains for only $T \approx 2\text{B}$  tokens. 
We therefore have not probed the $D \gg N$ regime at $T \sim D$. 
We have no specific reason to expect LAR to break out of its plateau in this scenario, but collecting more empirical evidence will be quite useful.
Third, our evidence is correlational: LAR decline coincides with overfitting, but we have not tested whether directly regularizing $\sigma_p$, $\sigma_q$, or $\operatorname{corr}(p,q)$ during training would actually prevent overfitting. 
Finally, we computed $q$ using a batch of fixed size in all pre-training experiments. Studying the behavior of LAR as a function of batch size and its relationship with generalization is an important future direction.

\section*{Acknowledgments}

We thank Michael Callahan and Philip Monk for help with training infrastructure for this project. We are also grateful to Adarsh Chaluvaraju, Devaansh Gupta, Yash Jain, Noam Shazeer, Somanshu Singla, and Kalina Slavkova for useful discussions, and to Yash Jain for comments on the draft.

\bibliography{bibliography}
\bibliographystyle{plainnat}

\appendix

\crefalias{section}{appendix} 
\crefalias{subsection}{appendix}

\section{Additional Experimental Details}
\subsection{Binary Operations For Grokking Experiments}
\label{app:tasks}

The 12 tasks from \citet{power2022grokking} are:
\begin{enumerate}
  \item $x + y \pmod{p}$
  \item $x - y \pmod{p}$
  \item $x / y \pmod{p}$ for $y \neq 0$
  \item $x/y \pmod{p}$ if $y$ odd, else $x - y \pmod{p}$
  \item $x^2 + y^2 \pmod{p}$
  \item $x^2 + xy + y^2 \pmod{p}$
  \item $x^2 + xy + y^2 + x \pmod{p}$
  \item $x^3 + xy \pmod{p}$
  \item $x^3 + xy^2 + y \pmod{p}$
  \item $x \cdot y$ for $x, y \in S_5$
  \item $x \cdot y \cdot x^{-1}$ for $x, y \in S_5$
  \item $x \cdot y \cdot x$ for $x, y \in S_5$
\end{enumerate}
Tasks 1--5 and 10--12 grok reliably. Tasks 6--9 do not grok with our experimental setup and our training budget (50{,}000 epochs with 50\% training fraction).

\subsection{Experimental details of large-scale experiments} \label{app:exp-details-large-scale}

 \paragraph{Architecture.} All models use Gemma-2 decoder blocks \citep{gemma2024} with $8$ transformer
  layers, embedding dimension $d_{\text{model}} = 4096$, MLP hidden dimension
  $d_{\text{ff}} = 16{,}384$ with a gated GeLU activation \citep{shazeer2020glu}, $32$ attention heads of head
  dimension $128$, and untied input/output embeddings over a vocabulary of $128{,}256$
  tokens (Llama-3 tokenizer \citep{llama3}). Both attention sub-layers within each Gemma-2 block are
  global (no sliding-window attention). Each attention and MLP sub-block is wrapped by RMSNorm \citep{zhang2019rmsnorm} both before \emph{and} after the sub-block (i.e., both pre-norm and post-norm are enabled, following the original Gemma-2 design \citep{gemma2024}); we do not apply QK-norm and do not soft-cap the final logits.                                                                     
                                                                                                                             
  \paragraph{Optimization.} We train for $10{,}000$ steps with a cosine learning-rate
  schedule \citep{loshchilov2017sgdr} preceded by linear warmup over the first $4\%$ of training ($400$ steps),
  and clip gradients at global norm $1.0$. For Adam \citep{kingma2015adam}, we use $\beta_1 = 0.95$, $\beta_2 = 0.999$, $\epsilon = 10^{-8}$, and no weight decay (we use the plain Adam optimizer rather than AdamW \citep{loshchilov2019decoupled}). For Muon, we use momentum $0.95$ with Nesterov updates, $5$ Newton--Schulz iterations for the orthogonalization step, and decoupled weight decay $\lambda = 0.1$ for all two-dimensional parameters including the unembedding matrix. We sweep the
  peak learning rate on the 100B-token dataset (in factor-of-two increments) and pick                                        
  the value with the lowest final validation loss for each optimizer; the resulting                                          
  optima are $\eta_{\text{Adam}} = 2.44 \times 10^{-4}$ and $\eta_{\text{Muon}} = 9.77                                       
  \times 10^{-4}$. These same learning rates are reused for the 200M, 400M, 800M, and 1.6B runs.                                                            
                                                                                                                             
  \paragraph{Dataset.} 

  The pretraining mixture is an internal Essential AI corpus consisting                                             
  predominantly of English web text with a smaller fraction of                                                             
  code.

\section{Additional Results on LAR and Related Quantities} \label{app:lar-derivations}

\subsection{Bounds on $\sum\limits_{i=1}^r p_i q_i$ and LAR}

Since $p$ and $q$ are distributions (over $r$ and $n$ elements respectively, with only the first $r$ terms of $q$ entering the sum):
\begin{align}
  \sum_{i=1}^r p_i q_i &\leq \max_j q_j \cdot \sum_{i=1}^r p_i = \max_j q_j \leq 1, \\
  \sum_{i=1}^r p_i q_i &\geq 0,
\end{align}
with equality in the lower bound iff $q_i = 0$ for all $i \leq r$ (i.e., $x$ is in the null space of $W$). The upper bound is achieved when $p$ and $q$ are both point masses on the same index. $\LAR$ is $-\infty$ at the lower bound and $\LAR = 1$ at the upper bound.

\subsection{$\sigma_p$ as $L_2$ distance to uniform distribution} \label{app:sigma-diffuse}

We derive the claim that $p$ is more diffuse iff $\sigma_p$ is small. Let $p$ be a distribution over $n$ elements. Expanding the squared $L_2$-norm from $p$ to the uniform distribution $u_i = 1/n$:
\[
  \|p - u\|^2 = \sum_i\!\left(p_i - \frac{1}{n}\right)^2 = \sum_i p_i^2 - \frac{1}{n},
\]
and therefore,
\[
  \sigma_p^2 = \frac{1}{n}\sum_i p_i^2 - \frac{1}{n^2} = \frac{1}{n}\|p - u\|^2.
\]
So $\sigma_p$ is proportional to the $L_2$ distance from $p$ to the uniform distribution, and is hence minimum when $p$ is uniform.

\subsection{Bounds on $\sigma_p$}

We can show that $\sigma_p$ is maximum when $p$ is concentrated on a single direction: $p_i = 1$ for a fixed $i$. To see this, note that transferring probability from $p_j$ to $p_k$ increases $\sigma_p$: suppose at least two components $p_j, p_k > 0$, then the change in $\sigma_p^2$ by transferring $\varepsilon > 0$ from $p_k$ to $p_j$ gives:
\[
  \Delta\!\left(\sum_i p_i^2\right) = (p_j+\varepsilon)^2 + (p_k-\varepsilon)^2 - p_j^2 - p_k^2 = 2\varepsilon(p_j - p_k) + 2\varepsilon^2 > 0
\]
Hence, transferring mass to a single component always increases $\sigma_p$. With $p_j = 1$, 
\[
\|p - u\|^2 = (1 - 1/n)^2 + (n-1)/n^2 = (n-1)/n,
\]
so $\sigma_{p,\max} = \sqrt{(n-1)/n^2} = \sqrt{n-1}/n$. In short,
\[
  \sigma_p \in \left[0,\; \frac{\sqrt{n-1}}{n}\right].
\]

\subsection{$\sigma_p$ and R\'enyi 2-entropy}
\label{app:sigma-p-renyi}

The R\'enyi 2-entropy of $p$ is defined as $H_2(p) = -\log_2 \sum_i p_i^2$, so $\sum_i p_i^2 = 2^{-H_2(p)}$. Substituting this into $\sigma_p^2 = \frac{1}{n}\sum_i p_i^2 - \frac{1}{n^2}$ gives
\[
  \sigma_p^2 \;=\; \frac{1}{n}\!\left(2^{-H_2(p)} - \frac{1}{n}\right).
\]
Hence, $\sigma_p$ is small iff $H_2(p)$ is close to its maximum, $\log_2 n$, which is attained when $p$ is uniform. The two quantities are monotonic functions of $\sum_i p_i^2$ and capture the same notion of diffuseness.

\subsection{LAR Dependence on Batch Size} \label{app:lar-batch}

 \begin{figure}[t]             
    \centering                                                                                                                        
    \begin{subfigure}[t]{0.48\columnwidth}                                                                                            
      \centering                                                                                                                      
      \includegraphics[width=\textwidth]{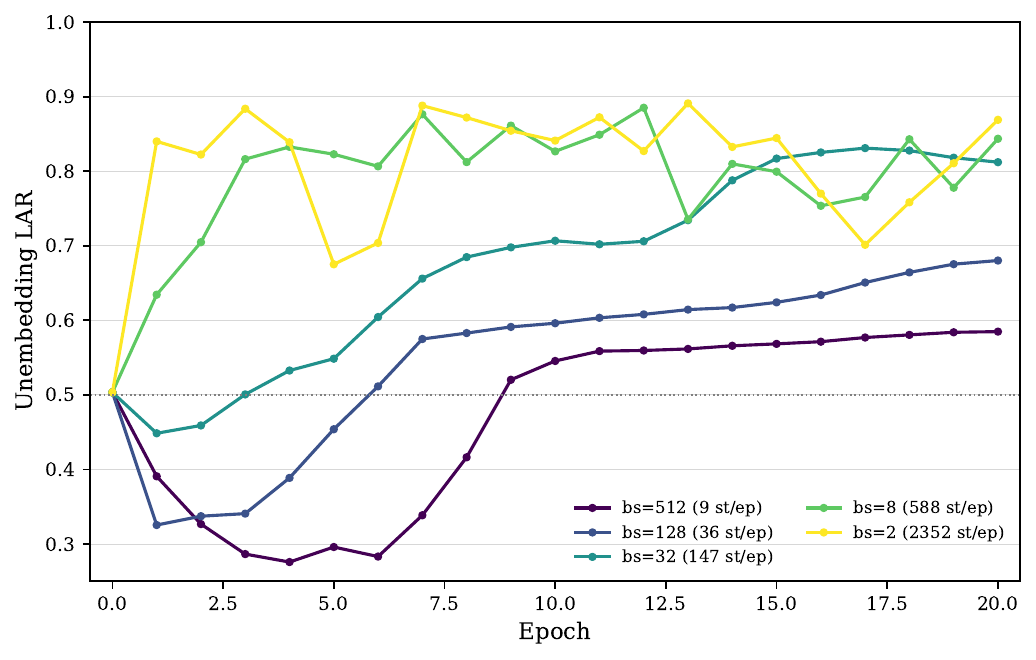}                                                            
      \caption{Unembedding LAR (y-axis) vs.\ training epochs (x-axis) for grokking on modular addition ($\bmod 97$). Legend entries give batch  size and the number of training steps per epoch.}
      \label{fig:early_lar_dip}
    \end{subfigure}                                                                                                                   
    \hfill        
    \begin{subfigure}[t]{0.48\columnwidth}
      \centering                                                                                                                      
      \includegraphics[width=\textwidth]{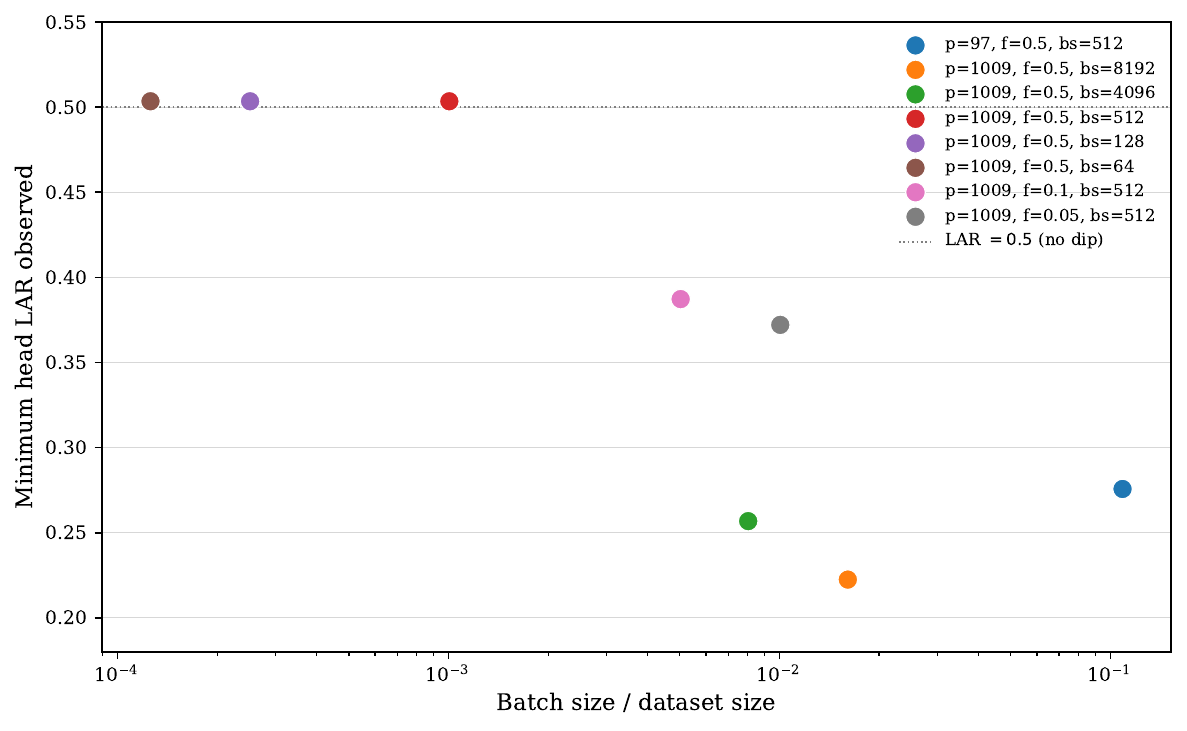}
      \caption{Minimum unembedding LAR across first 10 epochs (y-axis) and the ratio of batch size to training dataset size (x-axis) across grokking experiments on modular addition tasks with varying prime $p$, training data fraction and batch size.}  
      \label{fig:lar_min_vs_bs_over_n}
    \end{subfigure}                                                                                                                   
    \caption{Effect of the training batch size on unembedding LAR.}
    \label{fig:training_batch_lar}                                                                                             
  \end{figure}

Batch size affects LAR both directly, through $q$ (which is computed over the batch), and indirectly, through the optimization trajectory. 
More precisely, the batch size used to compute $q$ may be different from the optimization batch size; and we may study the effect of either of them on LAR.

\paragraph{Dependence of $\LAR$ on batch size through optimization.}  In grokking experiments of \cref{sec:grokking}, we compute $\LAR$ over the entire training dataset due to the small dataset sizes. 
Figure~\ref{fig:early_lar_dip} plots unembedding LAR over the first 20 epochs of training on    
  modular addition (for $p=97$) for 5 choices of batch size. 
  We observe an early dip in LAR at large batch sizes, which disappears when the batch size is small.
  We attribute this to the number of gradient updates per epoch: with few large 
  updates, each step perturbs weights enough to push them out of alignment with
  activations. This effect is transient, as $\LAR$ rises above $0.5$ after training for a few epochs.

To study whether this effect persists over a range of training dataset sizes and batch size, we vary training dataset fraction, batch size
  and the prime $p$ of modular addition in \cref{fig:lar_min_vs_bs_over_n}.
  We note that the minimum unembedding LAR observed in the first 10 training epochs decreases
  as the ratio of batch size to dataset size increases. This confirms that the presence of a few large gradient updates
  causes an early dip in unembedding LAR.
  In each case, training for more epochs pushed $\LAR$ above 0.5, confirming that the effect is only transient. 

\paragraph{Dependence of $q$ on batch size.} The activation distribution $q$ depends on the batch $B$ used to compute it. If the inputs $x \in B$ are drawn i.i.d.\ from a data distribution $\mathcal{D}$, then $q$ is an empirical estimate of the population quantity
\begin{equation}
  q_i^* \;=\; \frac{\mathbb{E}_{x\sim\mathcal{D}}\!\left[(v_i^\top x)^2\right]}{\mathbb{E}_{x\sim\mathcal{D}}\!\left[\|x\|^2\right]} \;=\; \frac{v_i^\top C\, v_i}{\operatorname{tr}(C)},
  \label{eq:q-star}
\end{equation}
where $C = \mathbb{E}_{x\sim\mathcal{D}}[x x^\top]$ is the population uncentered second-moment matrix. The batch version $q_i = \sum_{x \in B}(v_i^\top x)^2 / \sum_{x \in B}\|x\|^2$ replaces each expectation with its empirical average over $B$, and $q_i \to q_i^*$ as $|B| \to \infty$.

\section{Detailed Analysis of $p$ and $q$ During Pre-Training}
\label{app:pq-evolve}

In \cref{sec:large-scale} of the main text, we showed that $\LAR$ tracks the generalization gap in large-scale pre-training experiments. We also showed that the pre-overfit signature of overfitting lies mostly in the contribution from the top singular component, $L_1$. Here, we provide a more detailed analysis of $p$ and $q$ in \cref{app:pre-training-pq} and an alternative perspective on the relationship between $\LAR$ and overfitting in \cref{app:lar-spreads}.

\subsection{Detailed plots of $p$ and $q$} \label{app:pre-training-pq}

\begin{figure}[t]                      
    \centering                                                                                                
    \includegraphics[width=\textwidth]{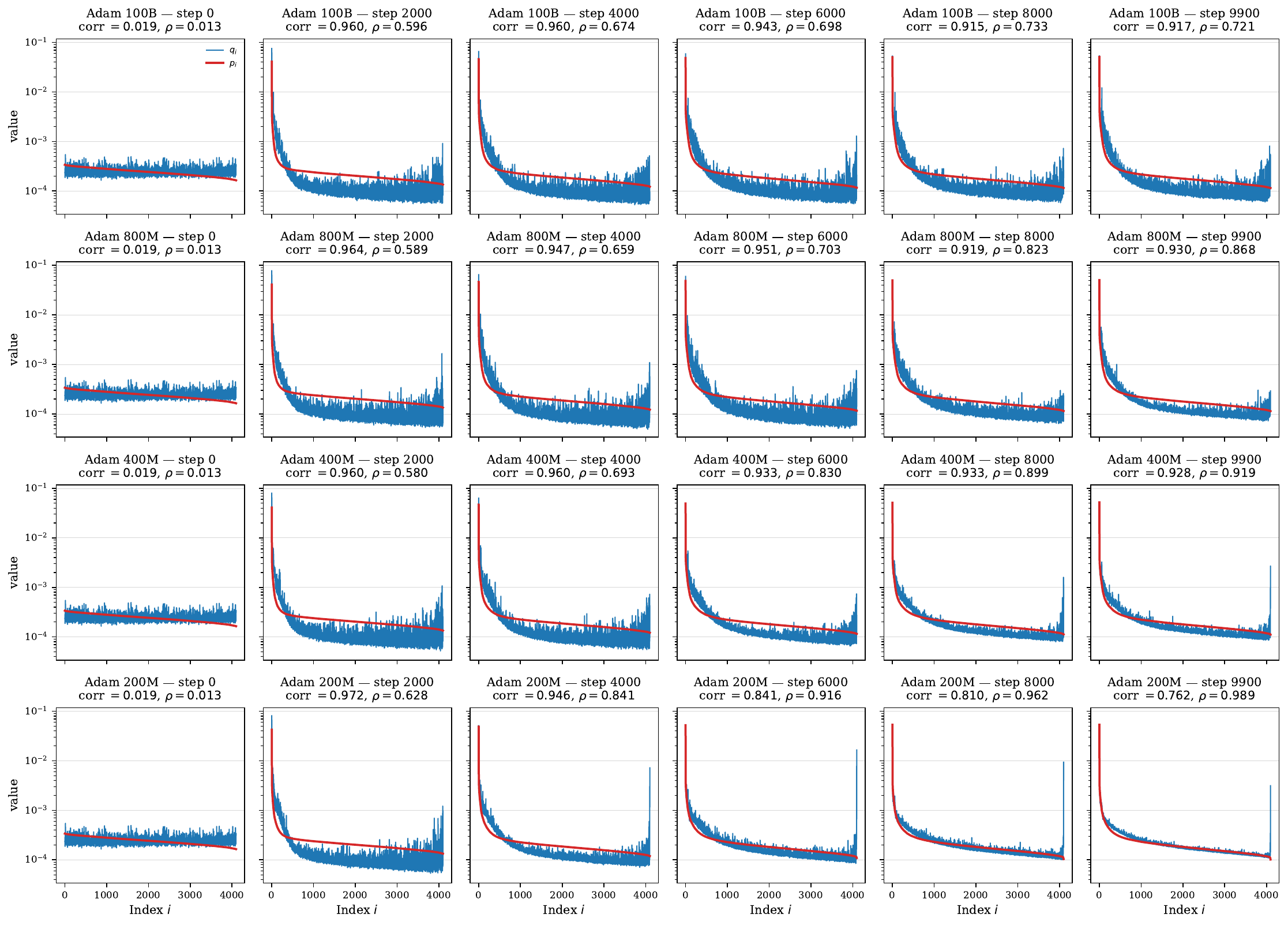}                            
    \caption{Weight distribution $p_i$ (red) and activation distribution $q_i$ (blue) against singular value indices $i$ of the       
  unembedding matrix $W$, for 3B models trained with Adam on datasets of 100B, 800M, 400M and 200M tokens. Each subplot title lists   
  Pearson correlation, $\corr(p, q)$, and Spearman's rank correlation, $\rho(p, q)$.}                                                 
    \label{fig:pq_adam_100b_vs_200m_per_step}                                                                                         
  \end{figure}

\begin{figure}[t]                                                                                                                   
    \centering    
    \includegraphics[width=\textwidth]{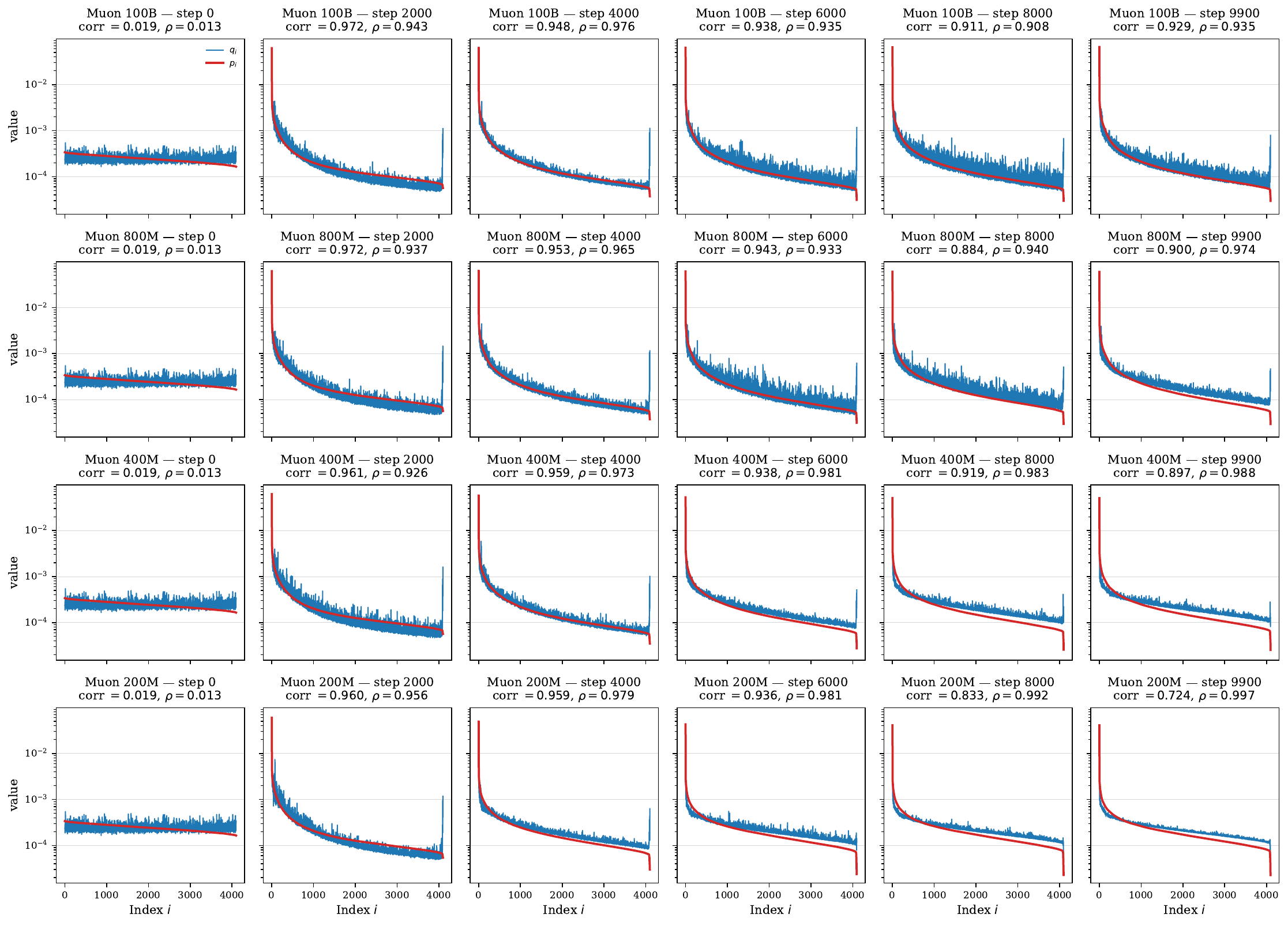}
    \caption{Weight distribution $p_i$ (red) and activation distribution $q_i$ (blue) against singular value indices $i$ of the       
  unembedding matrix $W$, for 3B models trained with Muon on datasets of 100B, 800M, 400M and 200M tokens. Each subplot title lists   
  Pearson correlation, $\corr(p, q)$, and Spearman's rank correlation, $\rho(p, q)$.}                                                 
    \label{fig:pq_muon_100b_vs_200m_per_step}                                                                                         
  \end{figure}

\begin{figure}[t]
    \centering
    \begin{subfigure}[t]{0.48\textwidth}
      \centering
      \includegraphics[width=\textwidth]{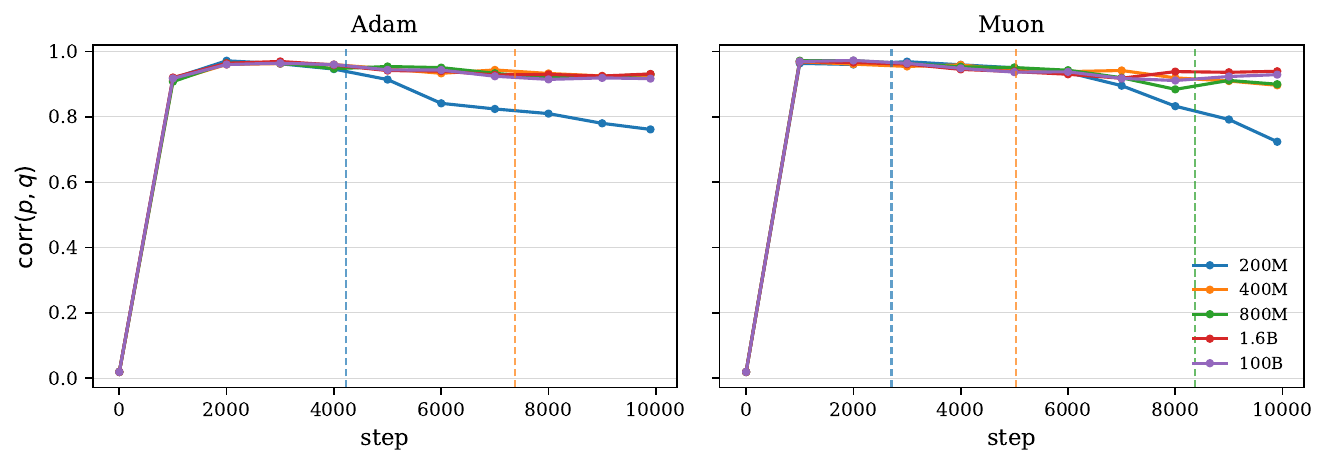}
      \caption{$\corr(p, q)$}
      \label{fig:pearson}
    \end{subfigure}
    \hfill
    \begin{subfigure}[t]{0.48\textwidth}
      \centering
      \includegraphics[width=\textwidth]{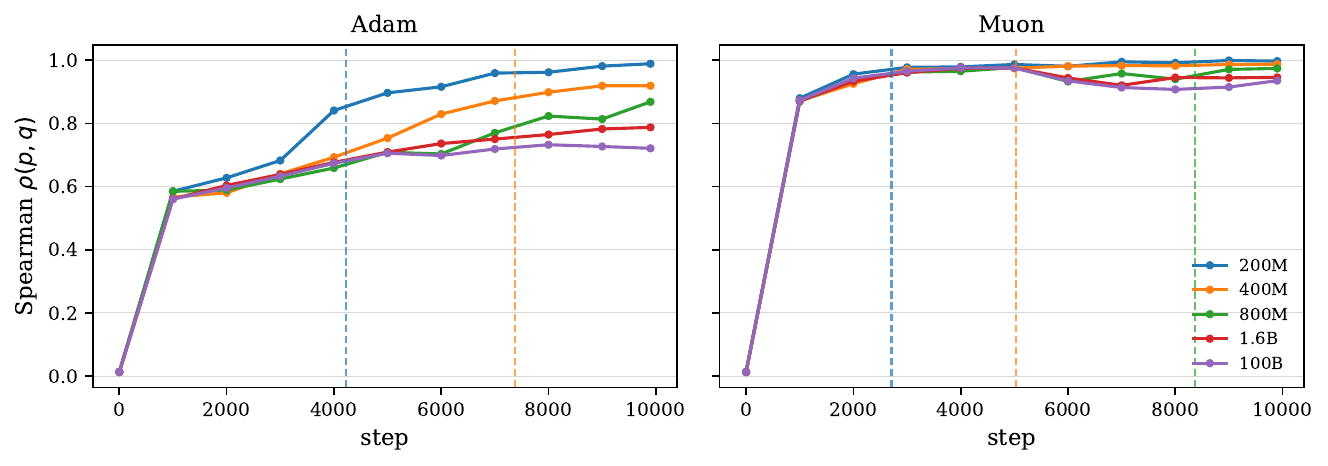}
      \caption{$\rho$}
      \label{fig:spearman}
    \end{subfigure}
    \vspace{0.5em}
    \begin{subfigure}[t]{0.48\textwidth}
      \centering
      \includegraphics[width=\textwidth]{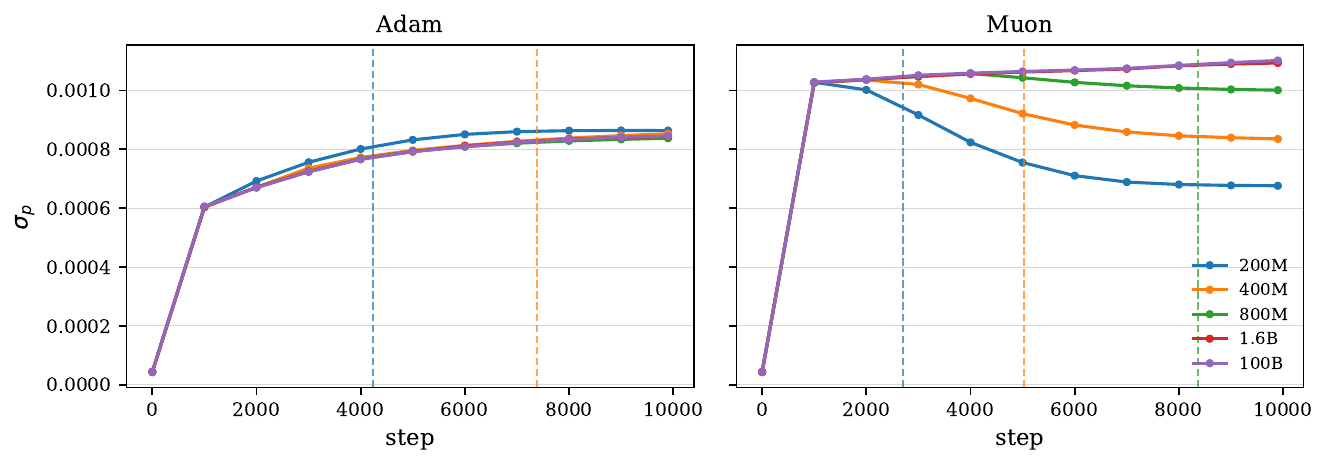}
      \caption{$\sigma_p$}
      \label{fig:sigmap}
    \end{subfigure}
    \hfill
    \begin{subfigure}[t]{0.48\textwidth}
      \centering
      \includegraphics[width=\textwidth]{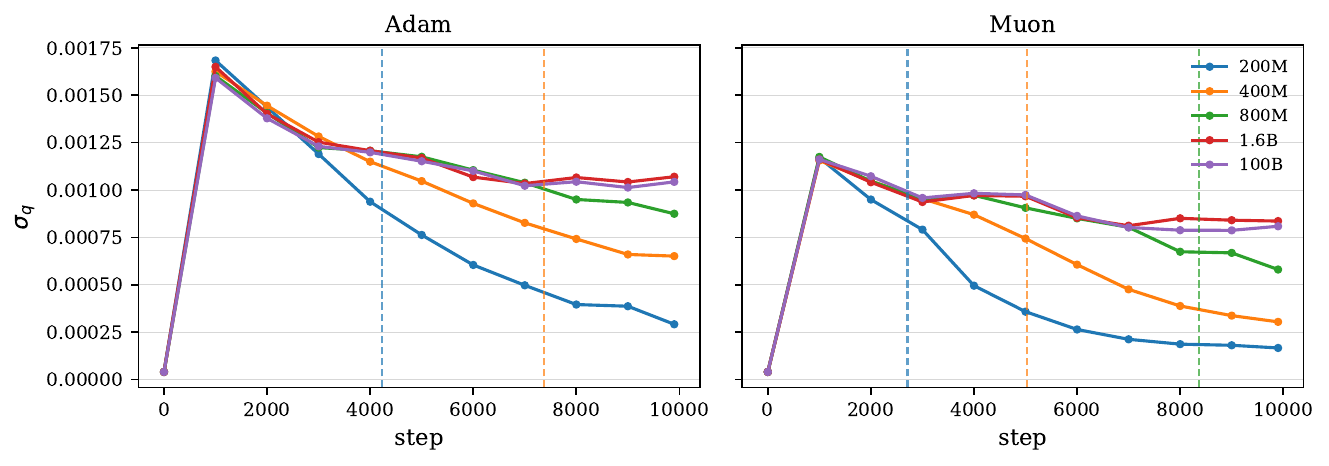}
      \caption{$\sigma_q$}
      \label{fig:sigmaq}
    \end{subfigure}

    \caption{Pearson correlation $\corr(p, q)$ (a), Spearman rank correlation $\rho$ (b), and standard deviations $\sigma_p$ (c) and $\sigma_q$ (d), as functions of training steps in 3B experiments. Different colors indicate dataset sizes; vertical dashed lines mark the onset of overfitting for the corresponding run.}
    \label{fig:sigmas-and-corr}
\end{figure}

In \cref{fig:pq_adam_100b_vs_200m_per_step} and \cref{fig:pq_muon_100b_vs_200m_per_step}, we give concrete examples of $p$ and $q$ for Adam and Muon experiments respectively. A row corresponds to a fixed dataset size: 100B, 800M, 400M or 200M, and a column represents a fixed training step: $0$, $2000$, $4000$, $6000$, $8000$, or $9900$.
\footnote{We skip the experiments with dataset size of 1.6B tokens as it is similar to the case of 100B dataset when training budget is 2B tokens.} 
Each subplot title contains Pearson correlation, $\corr(p, q)$, and Spearman rank correlation, $\rho(p, q)$, to capture linear and rank alignment between $p$ and $q$. We also plot Pearson and Spearman correlations every 1000 training steps for all 5 dataset sizes in \cref{fig:pearson} and \cref{fig:spearman} respectively.
  
We make the following observations:
\begin{itemize}
    \item \textbf{$p$ and $q$ at initialization.} At initialization, $p$ roughly follows the prediction from the Marchenko–Pastur distribution of the asymptotic behavior of singular values of random matrix. $q$ is isotropic with $q_i \approx 1/n = 1/4096 \approx 2 \times 10^{-4}$. 
    \item \textbf{$p$ during training.} After initialization, $p_i$ follows a decaying distribution across all training steps. The largest singular value is approximately 3 times larger than the next value, giving $p_1 \approx 9 p_2$. (See \cref{tab:sigma1_over_sigma2}.) Throughout training,
    $p_1 \approx 0.04-0.07$, or equivalently, $4-7\%$ of the weight's total energy. The smallest probability mass, $p_{4096}$, tends to be 1-2 orders of magnitude smaller compared to $p_2$: $p_{4096} \sim 10^{-4}$ for Adam and $p_{4096}  \sim 10^{-5}$ for Muon. 
    \item \textbf{$q$ during training.} Similar to $p$, $q$ also becomes more concentrated on the top singular directions but still remains more diffuse than $p$. The ratio $q_1 / q_2$ sits in the range $1.1$–$4.0$ across post-initialization checkpoints (median $\approx 2.3$; see \cref{tab:q1_over_q2}). $q_1$ plateaus around $0.04$ in the absence of overfitting, but overfitting causes a steady decline in its value. (See \cref{fig:q1_1x2_overfit} in the main text.)
    \item \textbf{Pearson Correlation between $p$ and $q$.}  At initialization, $\corr(p, q) \approx 0.02$ as  $q$ is isotropic and is uncorrelated with $p$. It rises sharply to above $0.9$ by step 1000 and remains there through training across all experiments, with the exception of 200M runs where prolonged training after overfitting causes a drop in the correlation. We leave a detailed study of this exception for future work.
    \item \textbf{Rank-alignment between $p$ and $q$.} Spearman correlation (\cref{fig:spearman}) shows that activation distribution and weight distribution exhibit larger rank-alignment with Muon as compared to Adam. In \cref{fig:pq_adam_100b_vs_200m_per_step} and \cref{fig:pq_muon_100b_vs_200m_per_step}, this reflects as $q$ taking larger values along the tail of $p$ in the case of Adam, but not in the case of Muon.

\end{itemize}

\begin{table}[t]
  \centering
  \caption{Ratio of the top two singular values $\sigma_1 / \sigma_2$ of the unembedding matrix at five training steps after initialization, for 3B pre-training experiments. The post-initialization value sits near 3 across all runs, so $p_1 \approx 9\, p_2$.}
  \label{tab:sigma1_over_sigma2}
  \begin{tabular}{l|ccccc|ccccc}
    \toprule
    & \multicolumn{5}{c|}{Adam} & \multicolumn{5}{c}{Muon} \\
    Dataset & 2000 & 4000 & 6000 & 8000 & 9900 & 2000 & 4000 & 6000 & 8000 & 9900 \\
    \midrule
    100B & 3.082 & 3.079 & 3.051 & 3.060 & 3.076 & 3.196 & 3.036 & 3.042 & 3.058 & 3.082 \\
    800M & 3.082 & 3.079 & 3.067 & 3.077 & 3.092 & 3.198 & 3.034 & 3.022 & 3.045 & 3.062 \\
    400M & 3.081 & 3.130 & 3.160 & 3.195 & 3.219 & 3.198 & 3.042 & 3.046 & 3.088 & 3.111 \\
    200M & 3.165 & 3.291 & 3.379 & 3.417 & 3.433 & 3.254 & 3.157 & 3.171 & 3.149 & 3.144 \\
    \bottomrule
  \end{tabular}
\end{table}

\begin{table}[t]
  \centering
  \caption{Ratio of the top two components of the activation distribution, $q_1 / q_2$, at five training steps after initialization, for our 3B pre-training experiments. Unlike $\sigma_1/\sigma_2 \approx 3$, this ratio is typically only $\sim 2$, indicating that $q$ is much more diffuse than $p$.}
  \label{tab:q1_over_q2}
  \begin{tabular}{l|ccccc|ccccc}
    \toprule
    & \multicolumn{5}{c|}{Adam} & \multicolumn{5}{c}{Muon} \\
    Dataset & 2000 & 4000 & 6000 & 8000 & 9900 & 2000 & 4000 & 6000 & 8000 & 9900 \\
    \midrule
    100B & 3.150 & 2.807 & 2.395 & 1.901 & 2.037 & 3.298 & 2.192 & 2.047 & 2.217 & 2.659 \\
    800M & 3.200 & 2.581 & 2.945 & 1.942 & 2.145 & 3.221 & 2.322 & 2.172 & 1.558 & 1.609 \\
    400M & 2.862 & 2.937 & 2.159 & 2.216 & 2.289 & 2.765 & 2.584 & 1.974 & 1.788 & 1.598 \\
    200M & 3.568 & 2.574 & 2.625 & 2.251 & 1.748 & 3.058 & 2.506 & 2.337 & 1.570 & 1.114 \\
    \bottomrule
  \end{tabular}
\end{table}

\subsection{LAR--generalization relationship through spreads in $p$ and $q$} \label{app:lar-spreads}

In \cref{sec:lar-definition}, we derived $\LAR$ in terms of the Pearson correlation, $\corr(p, q)$, and the standard deviations, $\sigma_p$ and $\sigma_q$ of $p$ and $q$, as
\[
\LAR = 1 + \frac{1}{2}\log_n\left(n \, \sigma_p \, \sigma_q \, \text{corr}(p, q) + \frac{1}{n} \right),
\]
where $\sigma_p^2 = \frac{1}{n} \sum\limits_{i=1}^n p_i^2 - \frac{1}{n^2}$ and $\sigma_q^2 = \frac{1}{n} \sum\limits_{i=1}^n q_i^2 - \frac{1}{n^2}$.

When $p_1 \gg p_i$ for $i\neq 1$, $\sigma_p \approx \frac{p_1}{\sqrt{n}}$. Similarly, when $q_1 \gg q_i$ for $i \neq 1$, $\sigma_q \approx \frac{q_1}{\sqrt{n}}$. As $\corr(p, q)$ stays approximately constant prior to overfitting, $\LAR$ is thus a proxy for $\sigma_p \sigma_q \approx \frac{p_1 q_1}{n} = n^{2L_1 - 3}$ in language-modeling pre-training experiments. The last equality follows from the definition of $L_1$.

In \cref{sec:large-scale}, we observed that $L_1$ contains most of the pre-overfit signature of overfitting. The behavior can thus equivalently be described in terms of the product of spreads of $p$ and $q$: $\sigma_p \sigma_q$. Indeed, $\sigma_p$ in \cref{fig:sigmap} has approximately the same shape as $p_1$ in \cref{fig:p1_1x2_overfit} and $\sigma_q$ in \cref{fig:sigmaq} has approximately the same shape as $q_1$ in \cref{fig:q1_1x2_overfit}. Furthermore, the pre-overfit signature of overfitting is visible in $\sigma_q$ for Adam and in both $\sigma_p$ and $\sigma_q$ for Muon. This gives a unified interpretation of the LAR--generalization behavior in both pre-training and grokking experiments in terms of measuring the spread of $p$ and $q$.

\section{Supplementary Results For Pre-training Experiments} \label{app:large-scale-additional}
\subsection{Comparison with other spectral measures} \label{app:other-spectral-measures}

\begin{figure}[t]
    \centering                                          
    \includegraphics[width=0.8\textwidth]{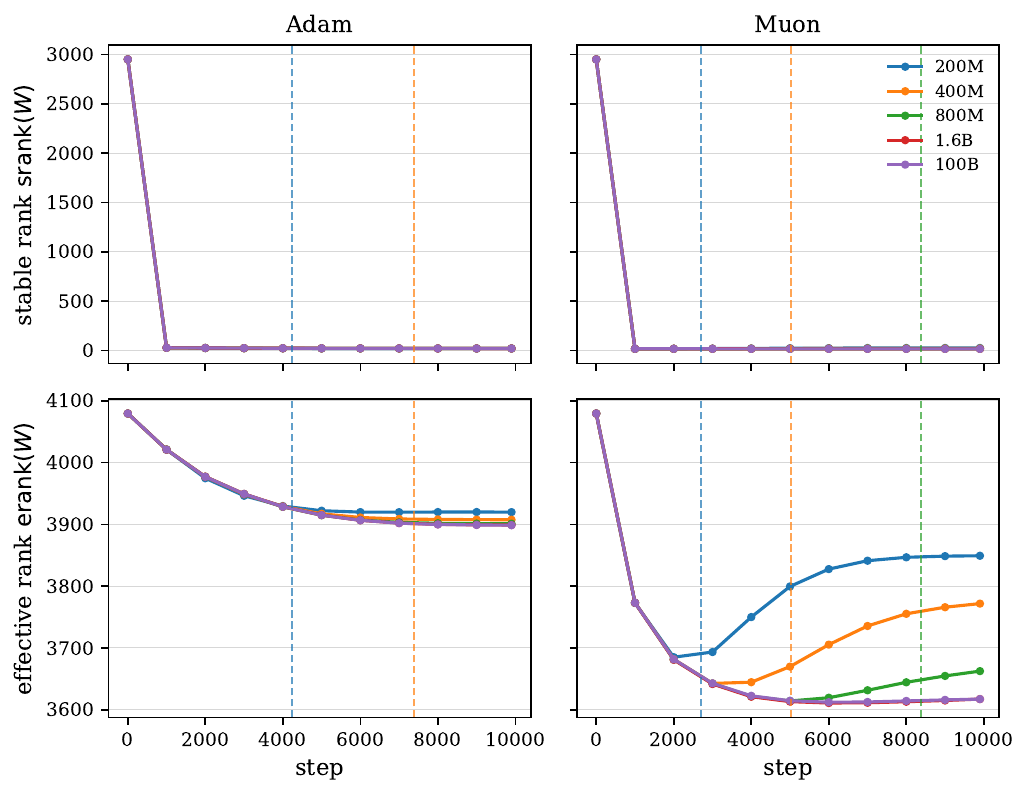}
    \caption{Stable rank (top) and effective rank (bottom) of the unembedding matrix as a function of training step, for the 3B models of \cref{sec:large-scale}. Vertical dashed lines mark the      
  overfitting onset for each run.}                                                              
    \label{fig:spectral-metrics}
  \end{figure}

Several spectral diagnostics of weight matrices have been proposed as                                                                                                                      
  indicators of generalization. We compare unembedding LAR against two of
  them -- stable rank and effective rank -- below, computing each every                                                                                                                      
  $1000$ training steps for the ten runs of \cref{sec:large-scale}                                                                                                                           
  (\cref{fig:spectral-metrics}). 

\paragraph{Stable rank.}
The stable rank of $W$ is defined as
\begin{equation}
    \mathrm{srank}(W) \;=\; \frac{\|W\|_F^2}{\|W\|_2^2} \;=\; \frac{1}{p_1}.
\end{equation}
\citet{neyshabur2018pac} (Theorem~1) prove a generalization bound for ReLU networks that is monotone in stable rank, predicting that lower stable rank corresponds to better generalization.
\citet{sanyal2020stable} used this result to introduce stable rank normalization as a
regularizer and show empirically that lowering stable rank improves
test accuracy across a range of vision classification and GAN tasks.

For 3B models of \cref{sec:large-scale}, $\mathrm{srank}(W)$ of the unembedding matrix
is $\sim 15-20$, as top singular values are much larger in magnitude
compared to the rest (cf. \cref{fig:spectral-metrics}, top row). Hence, it does not distinguish 
generalization from overfitting regime.

\paragraph{Effective rank.}
The effective rank of $W$ is defined in \citet{roy2007effective} as
\begin{equation}
    \label{eq:erank}
    \mathrm{erank}(W) \;=\; \exp\!\left(-\sum_{i=1}^{r} \tilde{p}_i \log \tilde{p}_i\right),
    \qquad \tilde{p}_i \;=\; \frac{s_i}{\sum_j s_j}.
\end{equation}
\citet{yunis2024spectral} track effective rank across grokking and pre-training tasks, 
and conclude that generalization corresponds to smaller values of effective rank.
\footnote{They define effective rank as the entropy itself (the logarithm of $\mathrm{erank}(W)$ in \cref{eq:erank}) and additionally divide by the matrix rank $R$. The qualitative direction is  
 the same.}

\cref{fig:spectral-metrics}, second row, shows that effective rank can distinguish overfitting models from
non-overfitting ones when trained with Muon. A similar distinction also exists in the case of Adam; however, the effect 
is quite small.

\subsection{Unembedding LAR dependence on learning rate schedule}

\begin{figure}[t]                                                                                                          
    \centering                                                                                                               
    \begin{subfigure}[t]{0.7\textwidth}                                                                                     
      \centering                                                                                                             
      \includegraphics[width=\textwidth]{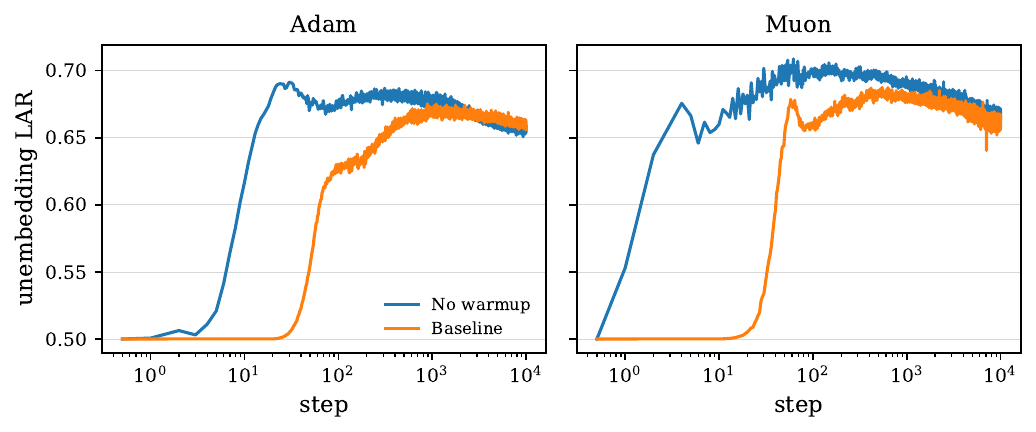}                                                   
      \caption{Effect of warmup. Without warmup, the LAR peak appears earlier, reflecting the larger learning rates used in  
  the initial optimization steps.}                                                                                           
      \label{fig:warmup_comparison}                                                                                          
    \end{subfigure}                                                                                                          
    \hfill        
    \begin{subfigure}[t]{0.7\textwidth}                                                                                     
      \centering  
      \includegraphics[width=\textwidth]{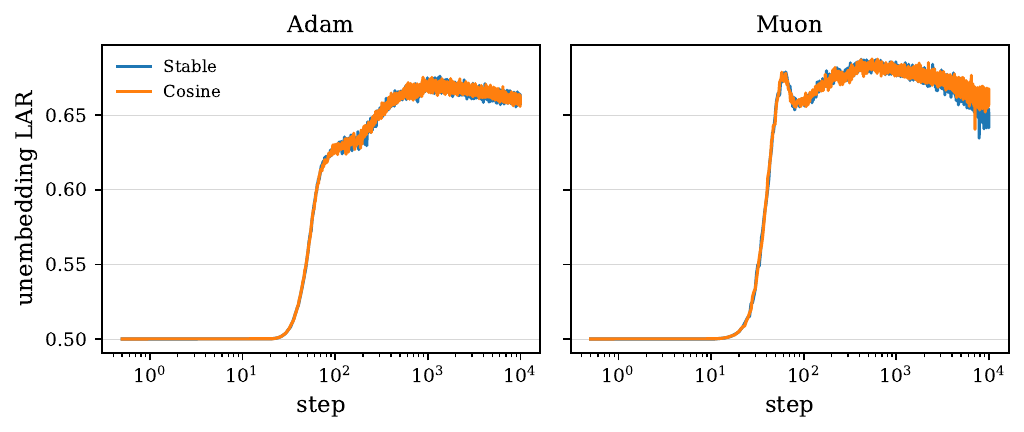}
      \caption{Effect of decay. Replacing cosine decay with a stable schedule (constant learning rate after warmup) has no   
  noticeable effect on LAR.}                                                                                                 
      \label{fig:schedule_comparison}                                                                                        
    \end{subfigure}                                                                                                          
    \caption{Effect of the learning rate schedule on unembedding LAR for the 100B dataset size.}
    \label{fig:lr_schedule_ablations}                                                                                        
  \end{figure} 

To understand the relationship between learning rate schedule and unembedding LAR, we run two additional experiments at   the 100B dataset size. \Cref{fig:warmup_comparison} compares the baseline (linear warmup over the first 4\% of steps)   
 against a no-warmup variant. The LAR peak appears earlier without warmup, which we attribute to the larger learning rates used in the early optimization steps. In \cref{fig:schedule_comparison}, we keep the 4\% warmup but hold the learning rate constant afterward (a stable 
  schedule) instead of applying cosine decay. This has no noticeable effect on unembedding LAR, even though the 
  stable-schedule model reaches a noticeably worse final loss than the cosine baseline (1.80 vs.\ 1.75).

\subsection{Effect of weight tying on unembedding LAR}
\begin{figure}[t]
    \centering                                          
    \includegraphics[width=0.7\textwidth]{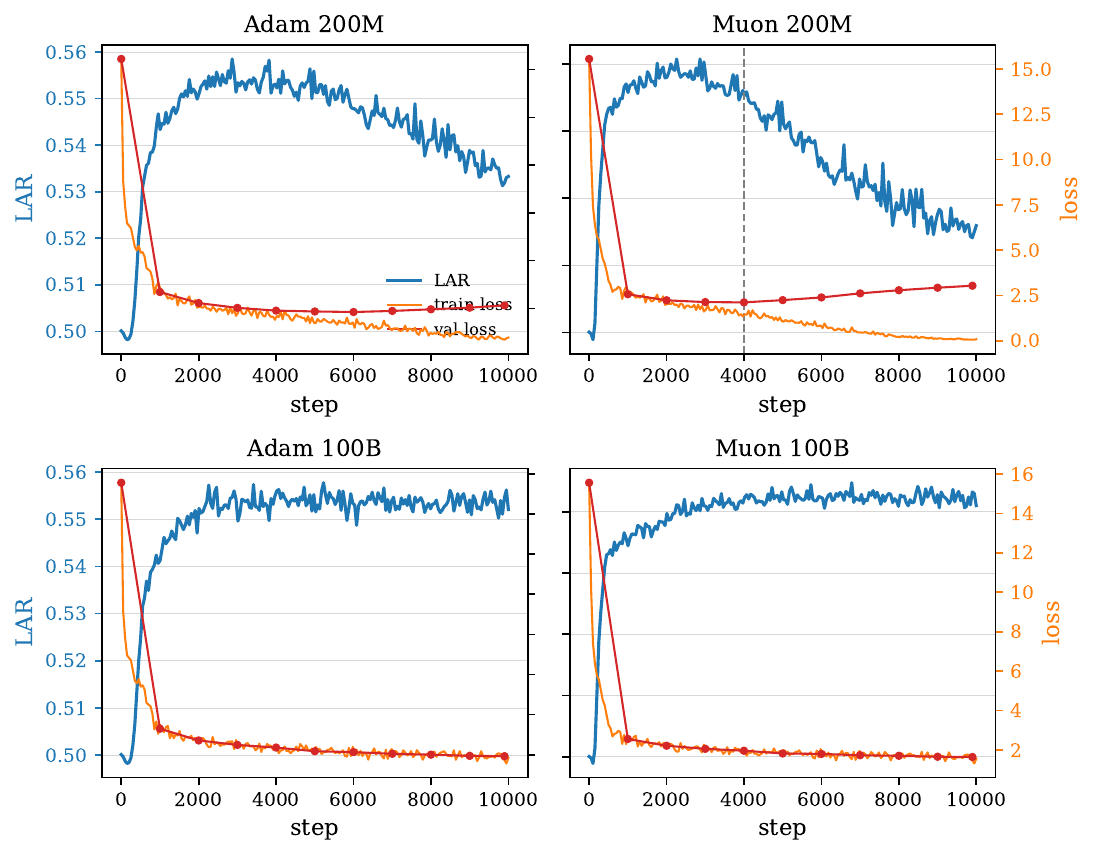}
    \caption{Training loss, validation loss, and unembedding log-alignment ratio (LAR) as functions of training steps for 2.5B models trained on 100B and 200M dataset sizes for 2B token budgets. All models used tied embeddings.}                                                              
    \label{fig:train_val_lar_tied_embeddings}
  \end{figure}

The models in \cref{sec:large-scale} use untied embeddings, while many modern LLMs use tied embeddings \citep{press2017using}. We construct tied-embedding variants of those models ($\sim$2.5B parameters) and train them on the 200M and 100B datasets for 2B tokens, matching the setup of \cref{sec:large-scale} 
 otherwise. \Cref{fig:train_val_lar_tied_embeddings} shows training loss, validation loss, and unembedding LAR; to save on evaluation cost, we log validation loss every 1000 steps here instead of every 20. The overall shape of LAR in the overfitting and non-overfitting regimes matches our 
 earlier findings, but peak LAR is noticeably smaller ($\sim$0.56 vs.\ $\sim$0.676--0.688 in \cref{sec:large-scale}). 

\subsection{Effect of weight decay on unembedding LAR}
\begin{figure}[t]
    \centering                                          
    \includegraphics[width=0.7\textwidth]{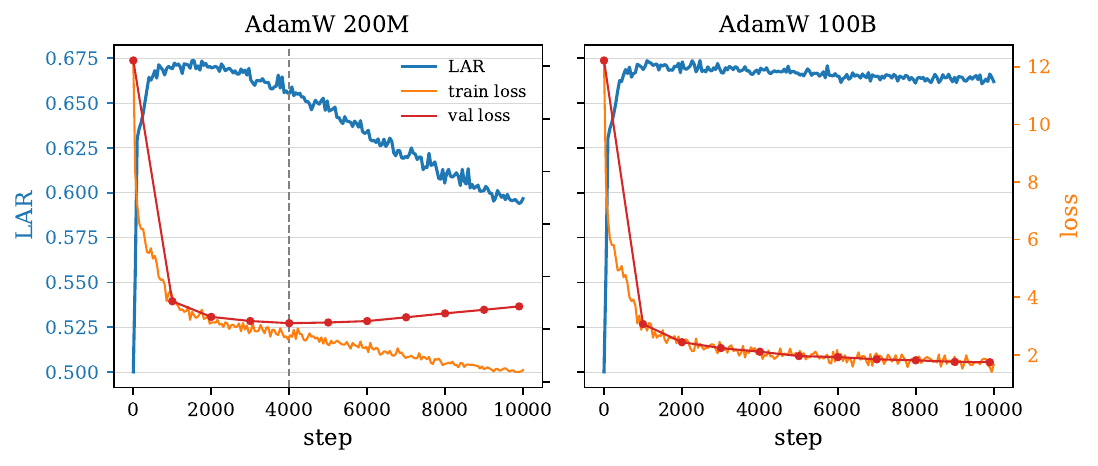}
    \caption{Training loss, validation loss, and unembedding log-alignment ratio (LAR) as functions of training steps for 3B models trained with AdamW on 100B and 200M dataset sizes for 2B token budgets.}                                                              
    \label{fig:train_val_lar_adamw}
  \end{figure}

The models trained with Adam in \cref{sec:large-scale} used no weight decay, while modern LLMs are typically trained with AdamW \citep{loshchilov2019decoupled}. To check that this choice does not affect our conclusions, we repeat two experiments of \cref{sec:large-scale} with AdamW and weight decay $0.1$. As shown in \cref{fig:train_val_lar_adamw}, the addition of weight decay has a negligible effect on the unembedding LAR trajectory. Validation loss is logged every 1000 steps instead of every 20 to reduce evaluation compute.
 
\subsection{LAR for other matrices in the network} \label{app:lar-other-matrices}

\begin{figure}[t]
    \centering                                          
    \includegraphics[width=0.9\textwidth]{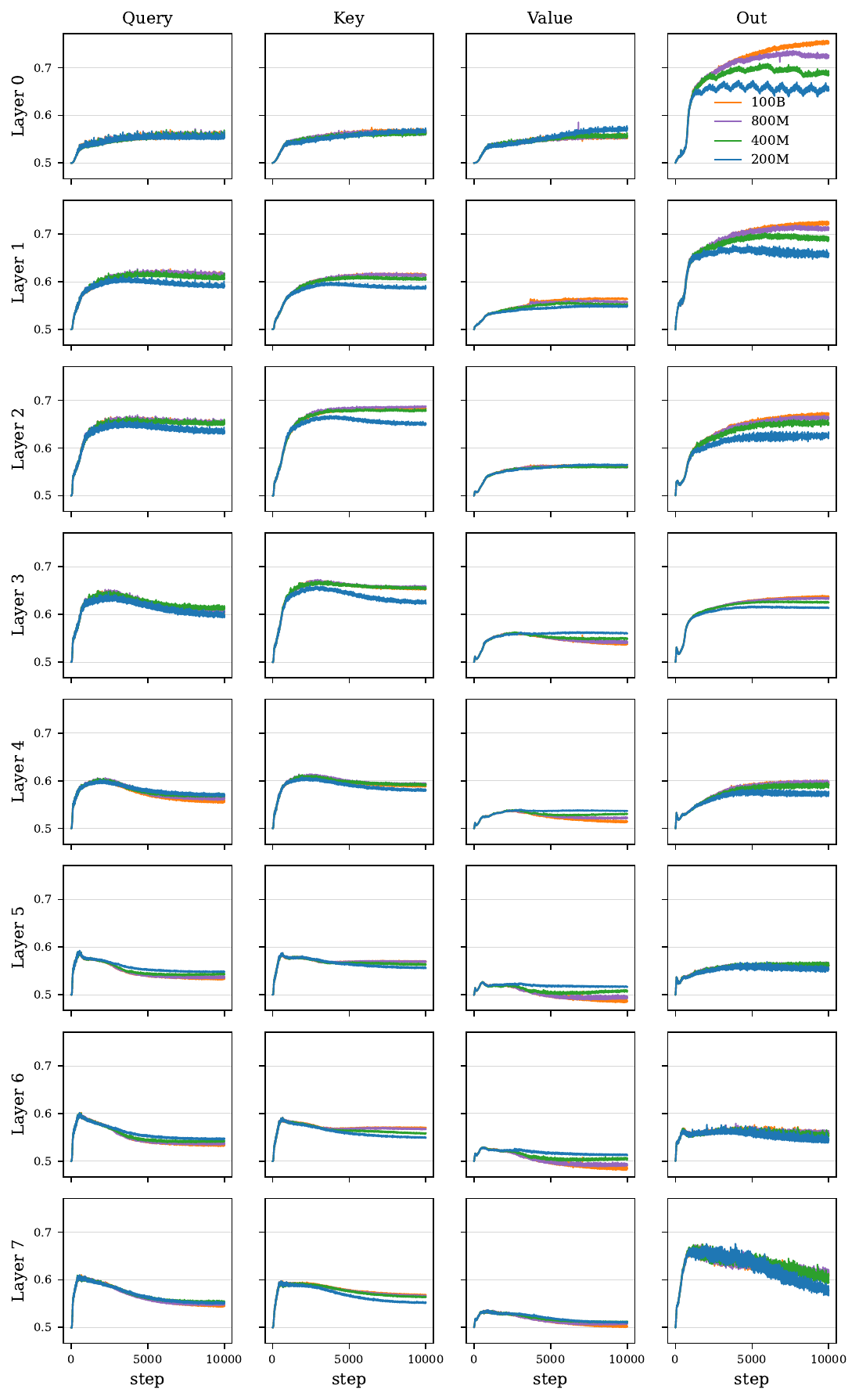}
    \caption{Log-alignment ratios of attention matrices in Adam runs. Each column shows one of the four attention projections ($W_q$, $W_k$, $W_v$,  
    $W_o$); each row shows one of the eight transformer layers.}                                                    
    \label{fig:attn_lar_per_layer_v13_adam_4sizes}
\end{figure}

\begin{figure}[t]
    \centering                                          
    \includegraphics[width=0.9\textwidth]{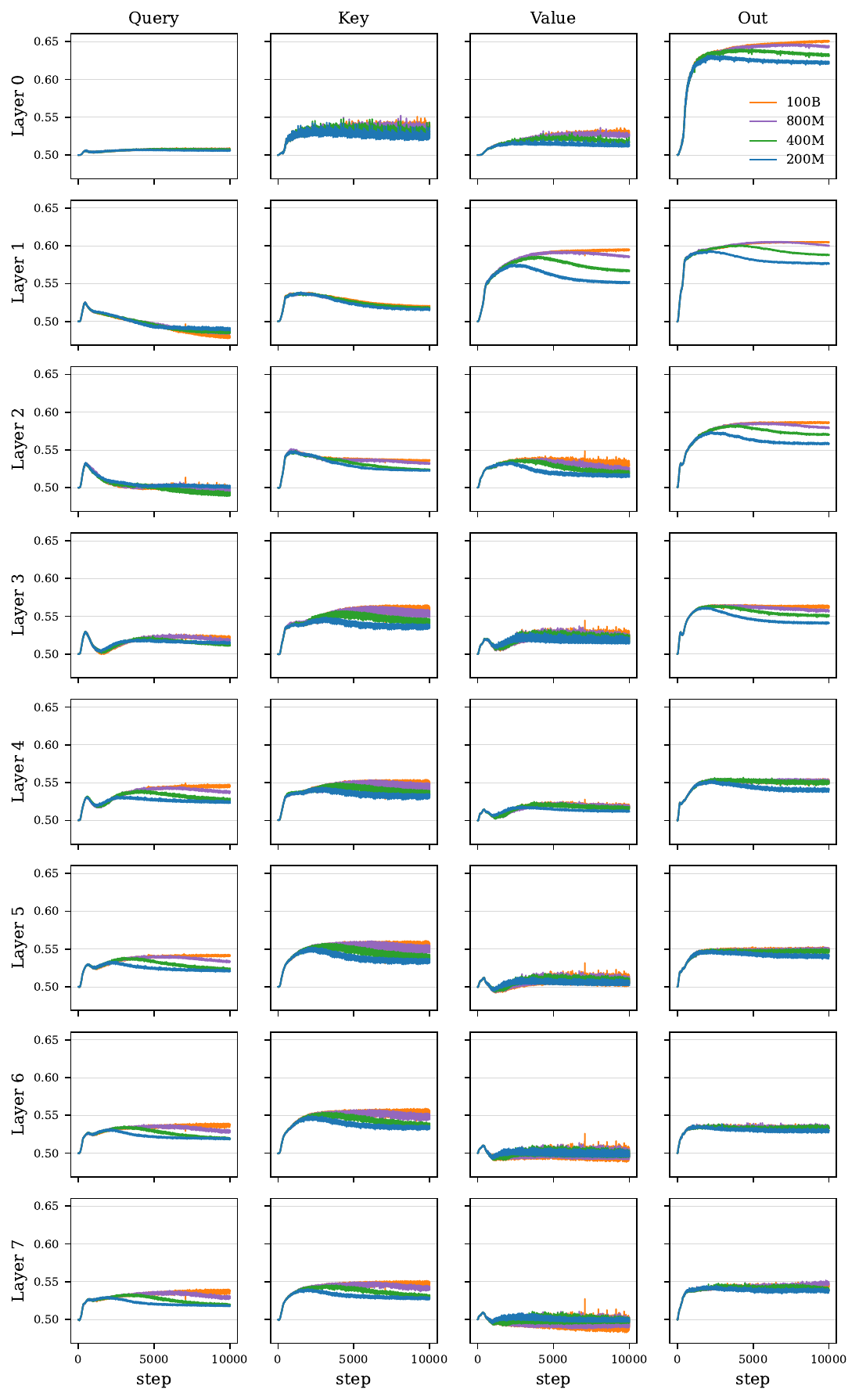}
    \caption{Log-alignment ratios of attention matrices in Muon runs. Each column shows one of the four attention projections ($W_q$, $W_k$, $W_v$,  
    $W_o$); each row shows one of the eight transformer layers.}                                                              
    \label{fig:attn_lar_per_layer_v13_muon_4sizes}
\end{figure}

\begin{figure}[t]
    \centering                                          
    \includegraphics[width=0.75\textwidth]{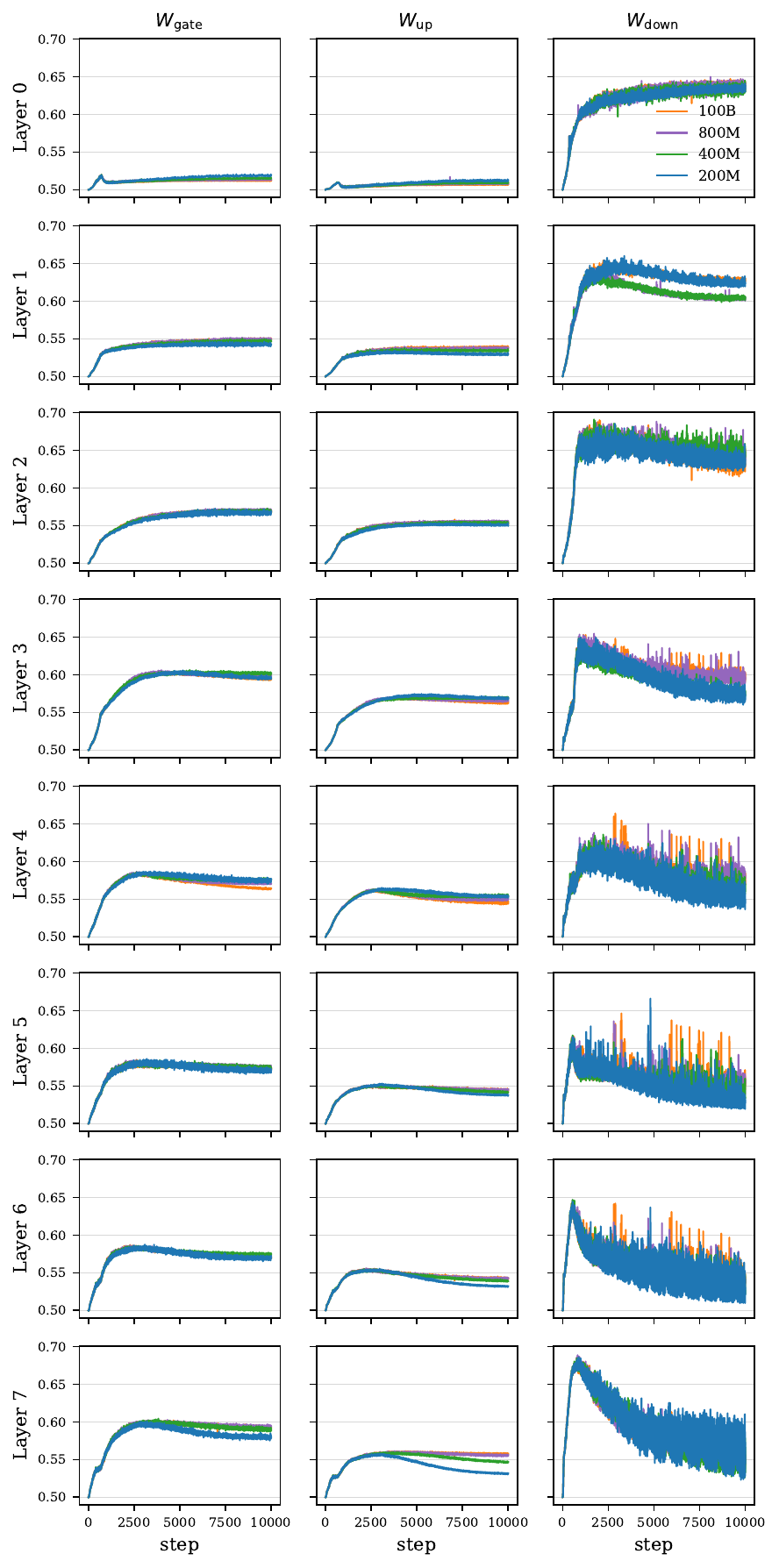}
    \caption{Log-alignment ratios of MLP matrices in Adam runs. Each column shows one of the three MLP projections ($W_\text{gate}$, $W_\text{up}$, $W_\text{down}$ ); each row shows one of the eight transformer layers.}                                                              
    \label{fig:mlp_lar_per_layer_v13_adam_4sizes}
  \end{figure}

\begin{figure}[t]
    \centering                                          
    \includegraphics[width=0.75\textwidth]{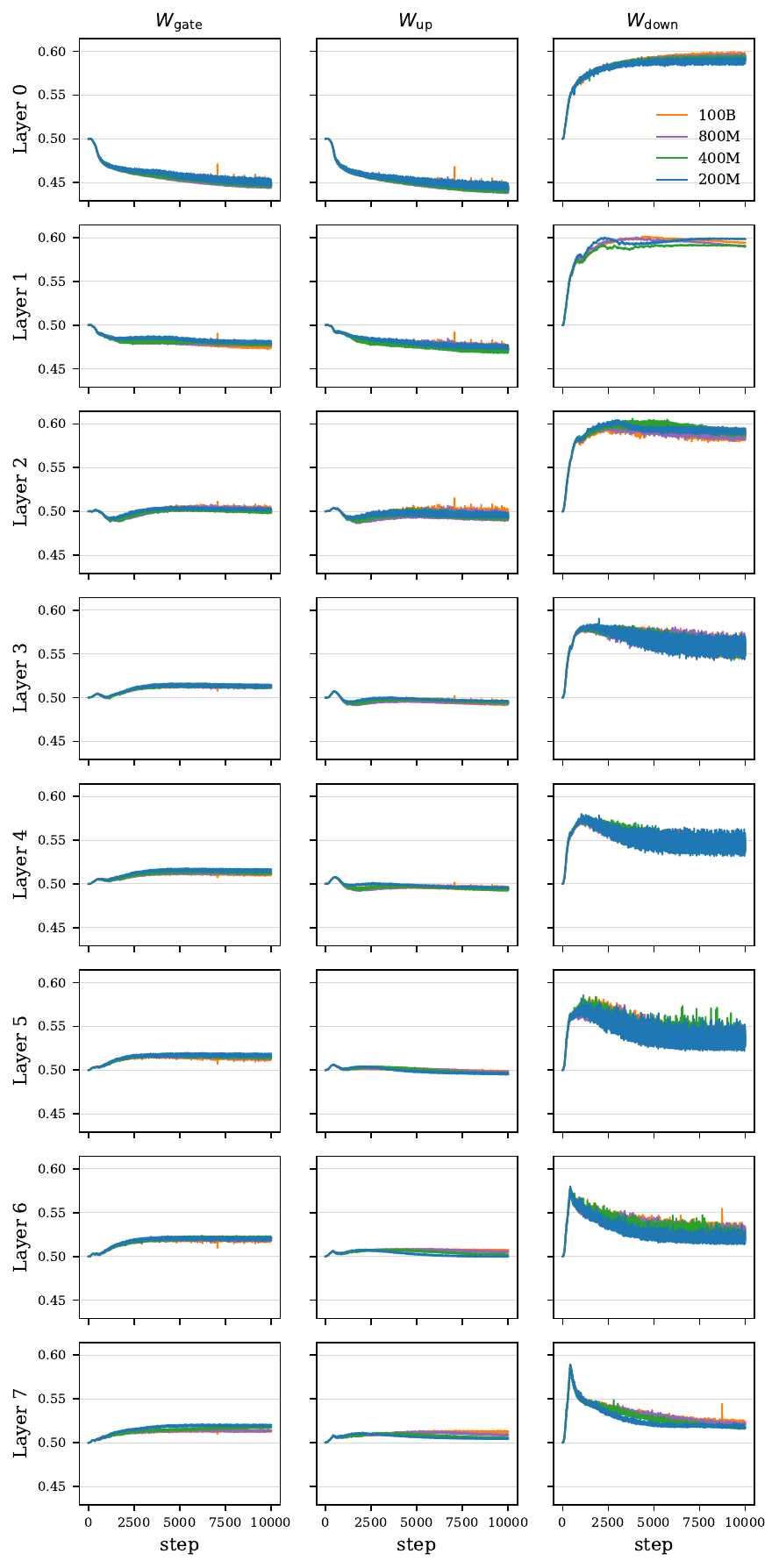}
    \caption{Log-alignment ratios of MLP matrices in Muon runs. Each column shows one of the three MLP projections ($W_\text{gate}$, $W_\text{up}$, $W_\text{down}$ ); each row shows one of the eight transformer layers.}                                                                  
    \label{fig:mlp_lar_per_layer_v13_muon_4sizes}
  \end{figure}

LAR is well-defined for any dense matrix in a neural network. For the models in \cref{sec:large-scale}, we plot LAR for each attention   
matrix in \cref{fig:attn_lar_per_layer_v13_adam_4sizes} (Adam) and \cref{fig:attn_lar_per_layer_v13_muon_4sizes} (Muon), and for each MLP
 matrix in \cref{fig:mlp_lar_per_layer_v13_adam_4sizes} (Adam) and \cref{fig:mlp_lar_per_layer_v13_muon_4sizes} (Muon). We omit the 1.6B 
 run since its behavior closely matches the 100B run within the 2B-token training budget.

 We make the following observations:

\begin{enumerate}
  \item LAR is approximately $\tfrac{1}{2}$ at initialization for every matrix in the network.

  \item Adam produces higher LAR than Muon at every internal matrix. The mean peak LAR under Adam exceeds that under Muon by roughly $0.05$ at both attention and MLP matrices, in every layer. We attribute this to Muon's spectral updates for internal matrices, which make the spectral distribution more uniform compared to Adam.
  (Equivalently, $\sigma_p$ is smaller for matrices optimized with Muon as compared to Adam.)
  \footnote{\citet{liu2025muon} report higher SVD entropy of weight matrices under Muon than AdamW, consistent with this observation.}

  \item The overfitting signature appears at a few other matrices in addition to the unembedding layer. The most notable is the divergence in LAR of the attention output matrices of earlier layers around overfitting epochs. 
  
\end{enumerate}

We leave a deeper analysis of LAR for other matrices to future work.

\section{Supplementary Results For Grokking Experiments}

\subsection{Additional Examples of $p$ and $q$} \label{app:grokking-pq}

\begin{figure}[t]
  \centering
  \begin{subfigure}[t]{\columnwidth}
    \centering
    \includegraphics[width=\columnwidth]{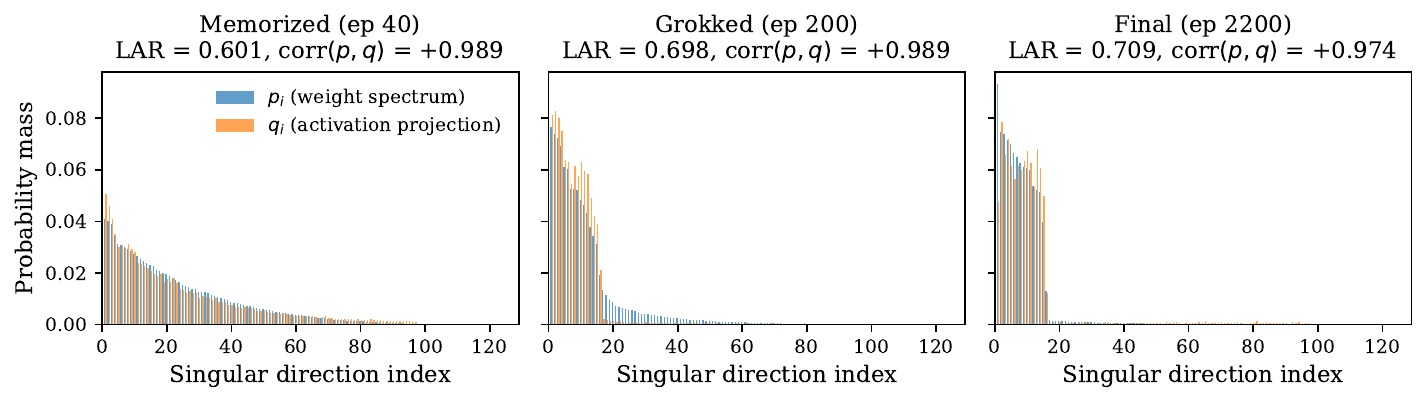}
    \caption{Modular addition, 50\% training data}
    \label{fig:pq_add50}
  \end{subfigure}
  \vspace{0.5em}
  \begin{subfigure}[t]{\columnwidth}
    \centering
    \includegraphics[width=\columnwidth]{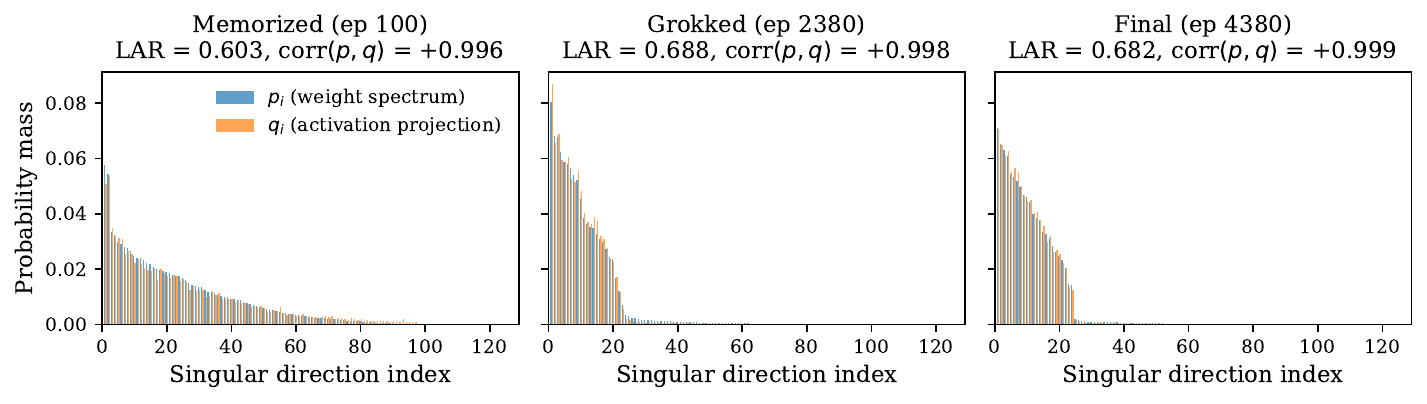}
    \caption{$x / y$ or $x - y$}
    \label{fig:pq_div_or_sub}
  \end{subfigure}
  \vspace{0.5em}
  \begin{subfigure}[t]{\columnwidth}
    \centering
    \includegraphics[width=\columnwidth]{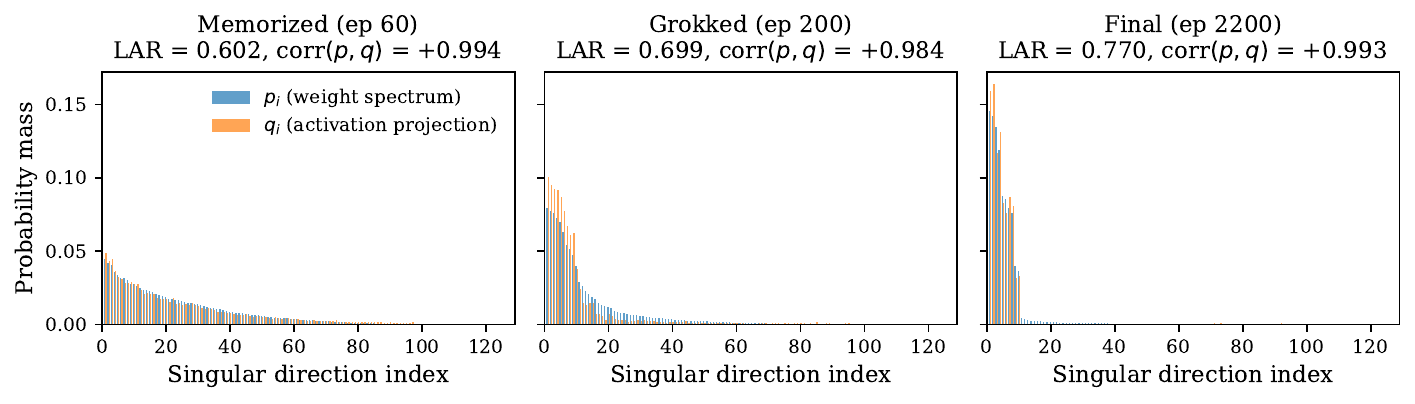}
    \caption{Modular subtraction}
    \label{fig:pq_sub}
  \end{subfigure}
  \vspace{0.5em}
  \begin{subfigure}[t]{\columnwidth}
    \centering
    \includegraphics[width=\columnwidth]{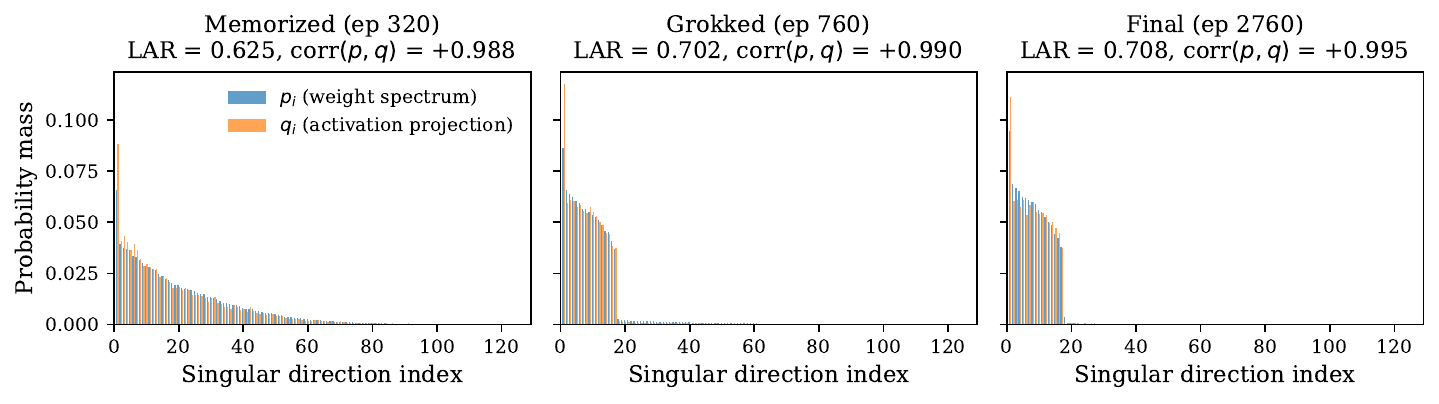}
    \caption{$x \cdot y \cdot x$ for $x, y \in S_5$}
    \label{fig:pq_s5_xyx}
  \end{subfigure}
  \caption{Weight distribution $p$ and activation distribution $q$ at memorization, grokking, and final points during training for tasks that grok.}
  \label{fig:pq_grokking_additional}
\end{figure}

\begin{figure}[t]
  \centering
  \begin{subfigure}[t]{0.7\columnwidth}
    \centering
    \includegraphics[width=\textwidth]{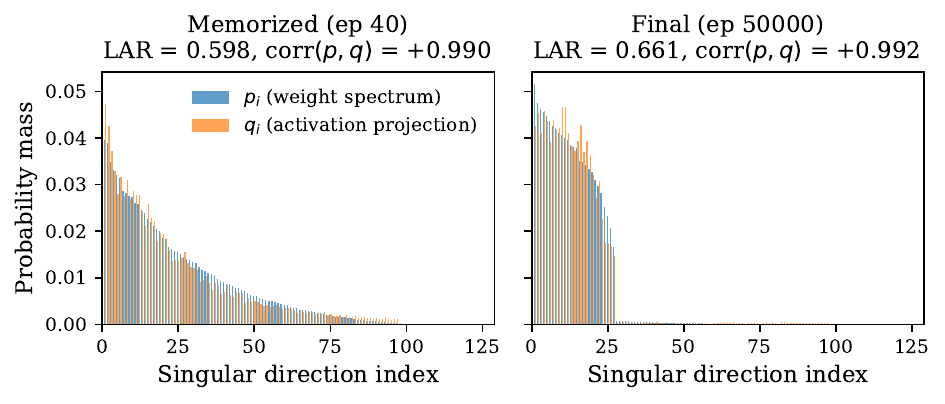}
    \caption{Modular addition, 50\% training data}
    \label{fig:grokking_frac_a}
  \end{subfigure}
  \vspace{0.5em}
  \begin{subfigure}[t]{0.7\columnwidth}
    \centering
    \includegraphics[width=\textwidth]{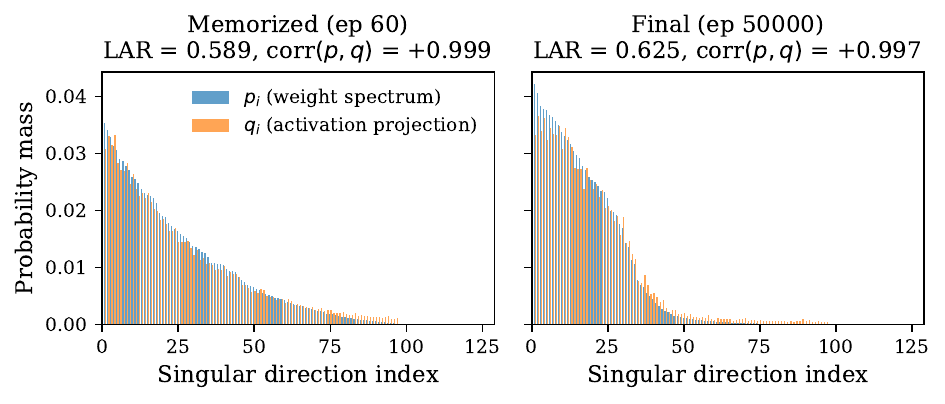}
    \caption{$x / y$ or $x - y$}
    \label{fig:grokking_frac_b}
  \end{subfigure}
  \caption{Weight distribution $p$ and activation distribution $q$ at memorization and final points during training for tasks that do not grok.}
  \label{fig:pq_no_grokking_additional}
\end{figure}

In \cref{fig:grokking-pq-combined}, we presented a few examples of $p$ and $q$ at different steps during training. Here, we present additional examples for grokking models (\cref{fig:pq_grokking_additional}) and non-grokking models (\cref{fig:pq_no_grokking_additional}). The assumptions of property 3 from \cref{sec:lar-properties}, i.e. $p_i \approx \frac{1}{k}$ for $i \leq k$ and $q_i = 0$ for $i > k$, hold true to varying degree at final checkpoints. $p$ and $q$ are invariably concentrated on the same $k$ directions, but the uniformity condition on $p$ is task-dependent. As exhibited in \cref{fig:kpred_v_k95_tasks}, $k \approx n^{2(1-\LAR)}$ approximates $k_{95}$ quite well, nevertheless. It is also clear that $p$ and $q$ are more diffuse at memorization and more concentrated at grokking and final checkpoints in all cases.

\subsection{Additional Plots of Unembedding LAR and Validation Accuracy} \label{app:lar-with-other-metrics}

\begin{figure}[t]
    \centering
    \begin{subfigure}{\columnwidth}
      \centering
      \includegraphics[width=0.7\linewidth]{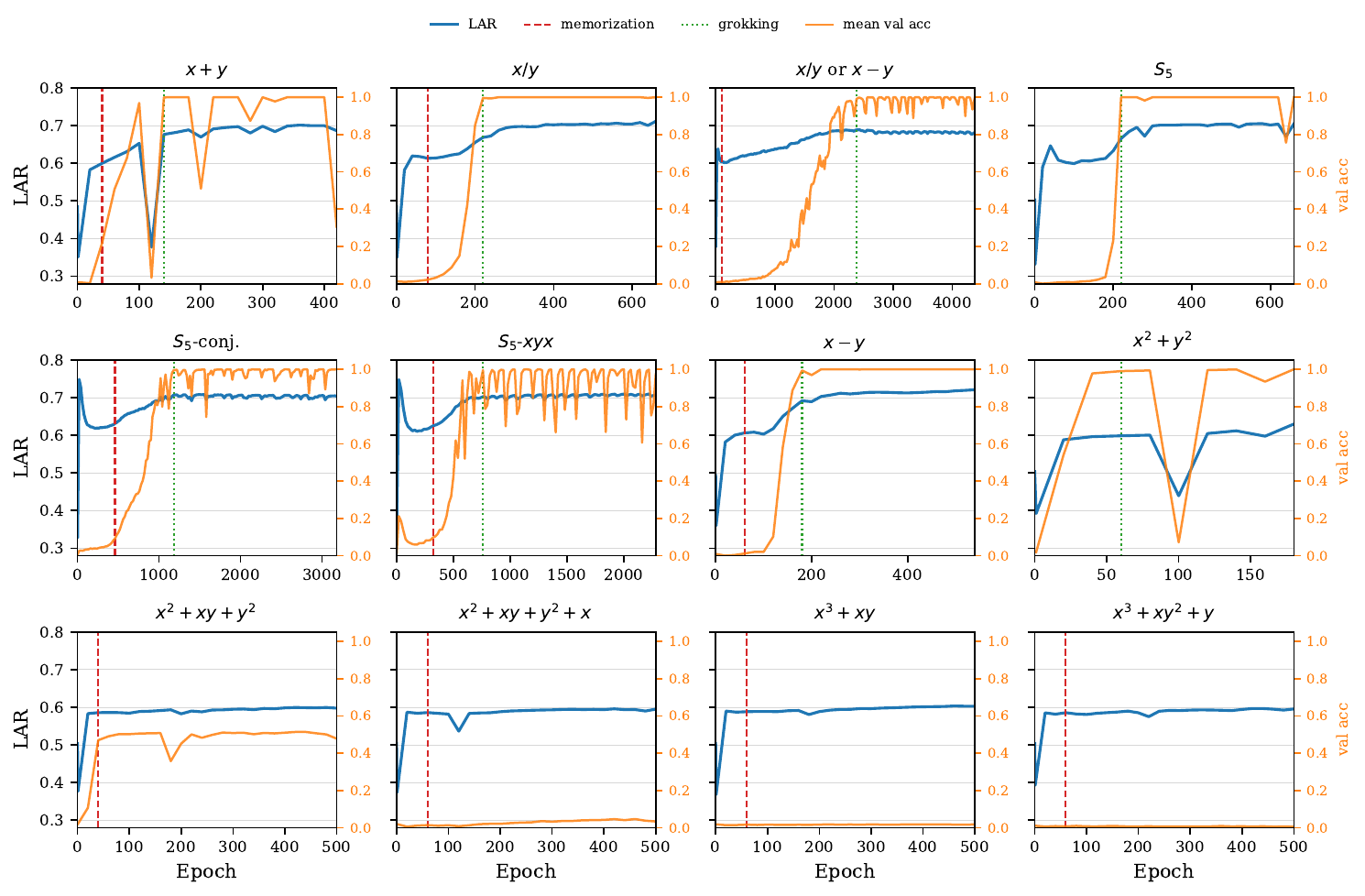}
      \caption{Seed 1}\label{fig:grokking-12-tasks-seed1}
    \end{subfigure}\\[0.5em]
    \begin{subfigure}{\columnwidth}
      \centering
      \includegraphics[width=0.7\linewidth]{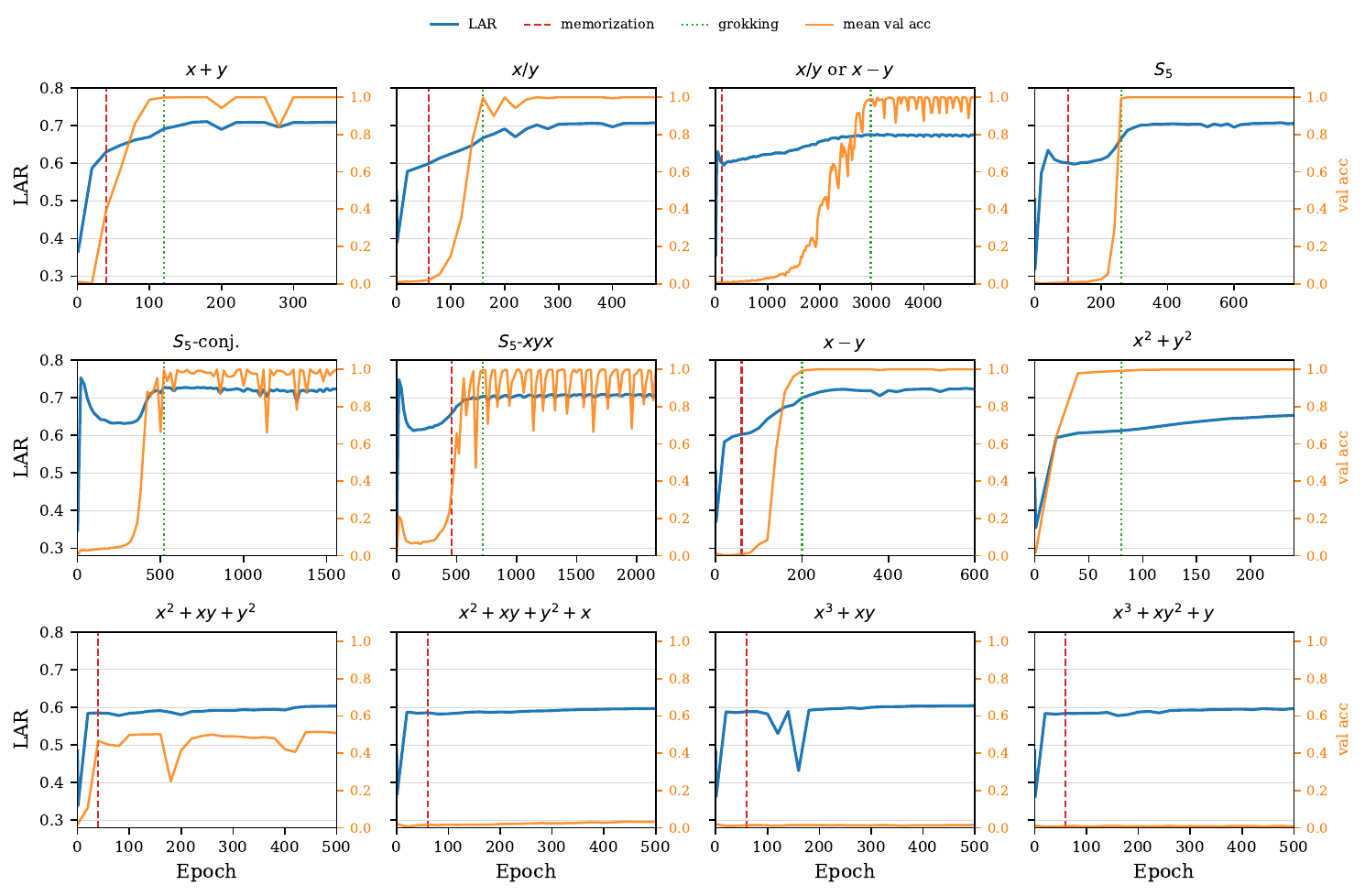}
      \caption{Seed 2}\label{fig:grokking-12-tasks-seed2}
    \end{subfigure}\\[0.5em]
    \begin{subfigure}{\columnwidth}
      \centering
      \includegraphics[width=0.7\linewidth]{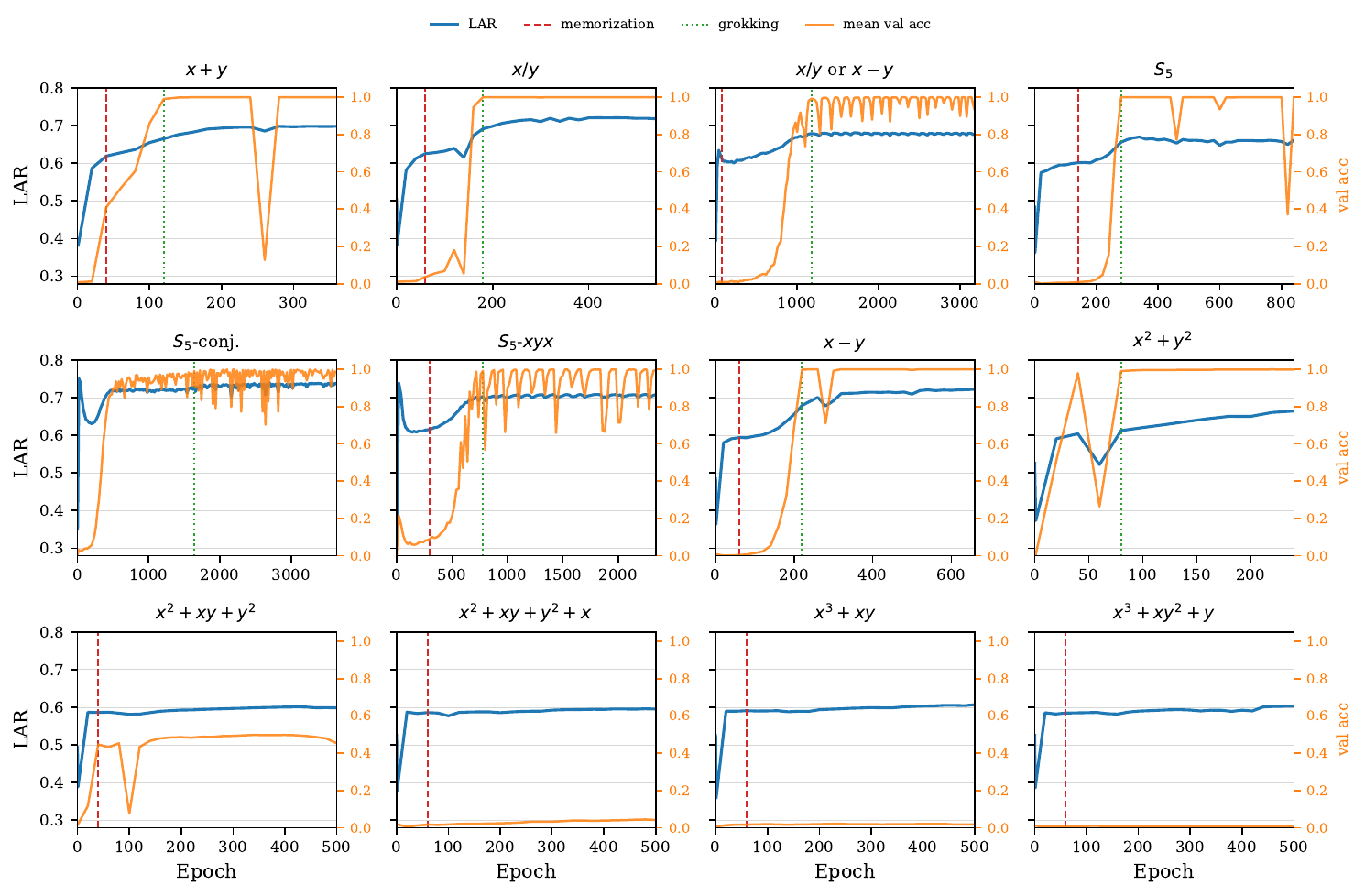}
      \caption{Seed 3}\label{fig:grokking-12-tasks-seed3}
    \end{subfigure}
    \caption{Unembedding LAR and validation accuracy across the 12 binary tasks of \cref{sec:grokking} for three representative seeds. The memorization-to-grokking pattern seen in the mean across seeds (\cref{fig:grokking-12-tasks}) is also apparent in individual runs.}
    \label{fig:grokking-lar-12-tasks-seeds}
  \end{figure}

\begin{figure}[t]
    \centering
    \begin{subfigure}{\columnwidth}
      \centering
      \includegraphics[width=\linewidth]{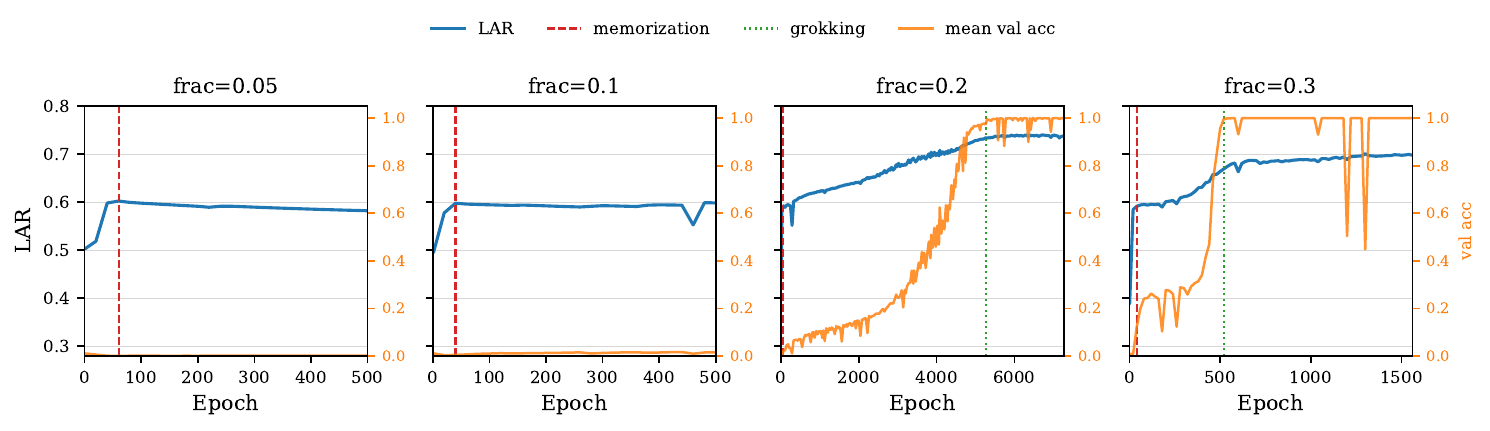}
      \caption{Seed 1}\label{fig:grokking-frac-tasks-seed1}
    \end{subfigure}\\[0.5em]
    \begin{subfigure}{\columnwidth}
      \centering
      \includegraphics[width=\linewidth]{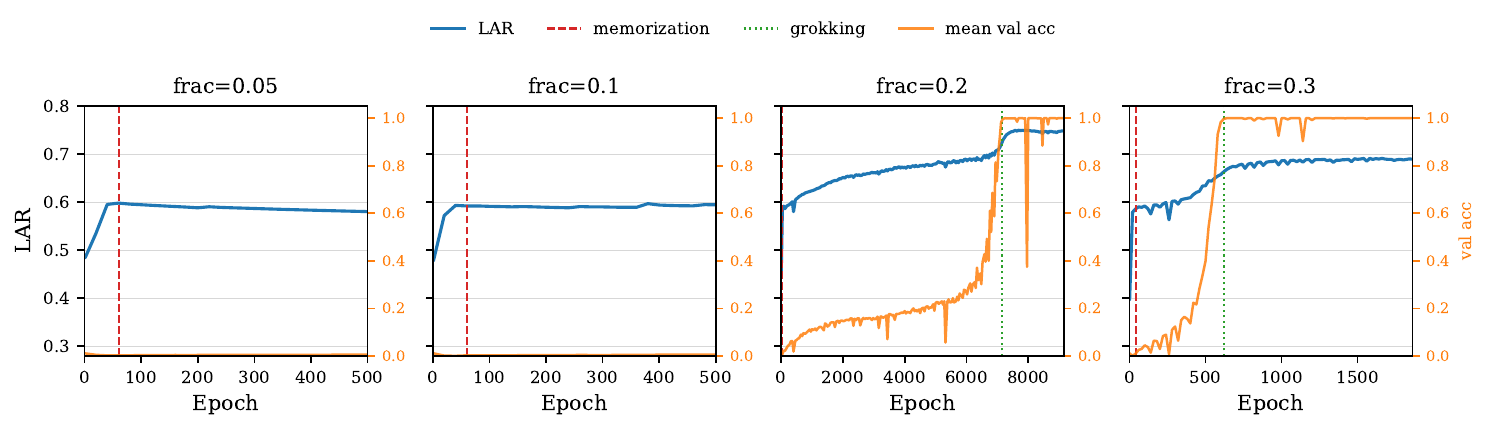}
      \caption{Seed 2}\label{fig:grokking-frac-tasks-seed2}
    \end{subfigure}\\[0.5em]
    \begin{subfigure}{\columnwidth}
      \centering
      \includegraphics[width=\linewidth]{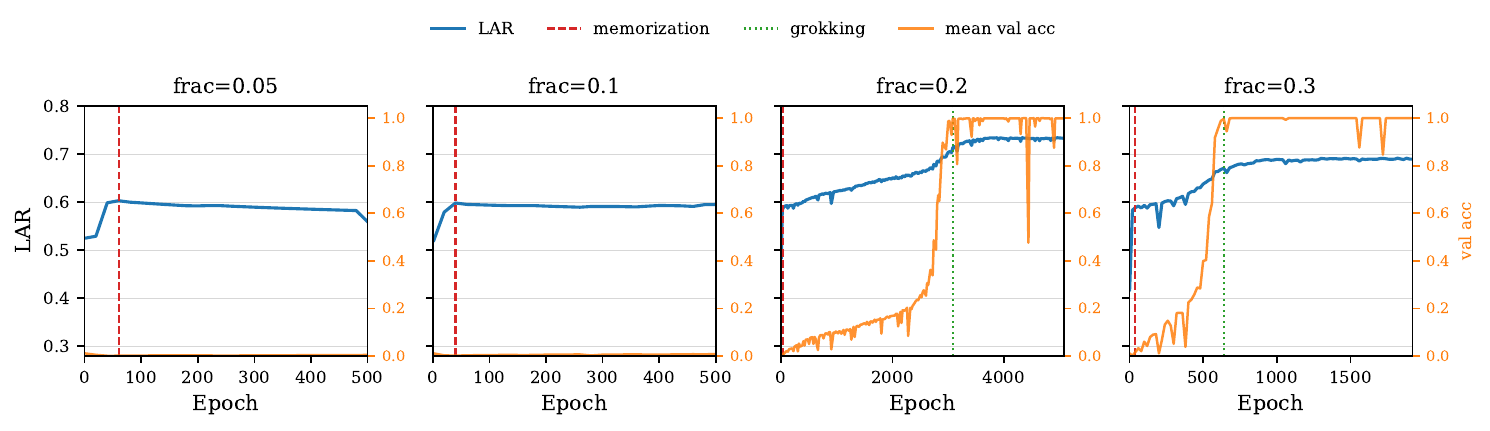}
      \caption{Seed 3}\label{fig:grokking-frac-tasks-seed3}
    \end{subfigure}
    \caption{Unembedding LAR and validation accuracy across the training-dataset-fraction-sweep experiments of \cref{sec:grokking} for three representative seeds. The memorization-to-grokking pattern seen in the mean across seeds (\cref{fig:grokking_dataset_fraction}) is also apparent in individual runs.}
    \label{fig:grokking-lar-frac-tasks-seeds}
  \end{figure}

In \cref{fig:grokking-lar-combined} of the main text, we showed mean unembedding LAR and mean validation accuracy across different seeds for the two sets of grokking experiments. Here, we provide these plots for individual seeds in \cref{fig:grokking-lar-12-tasks-seeds} and \cref{fig:grokking-lar-frac-tasks-seeds} respectively. 

\begin{figure}[t]
  \centering
  \begin{subfigure}[t]{\columnwidth}
    \centering
    \includegraphics[width=\columnwidth]{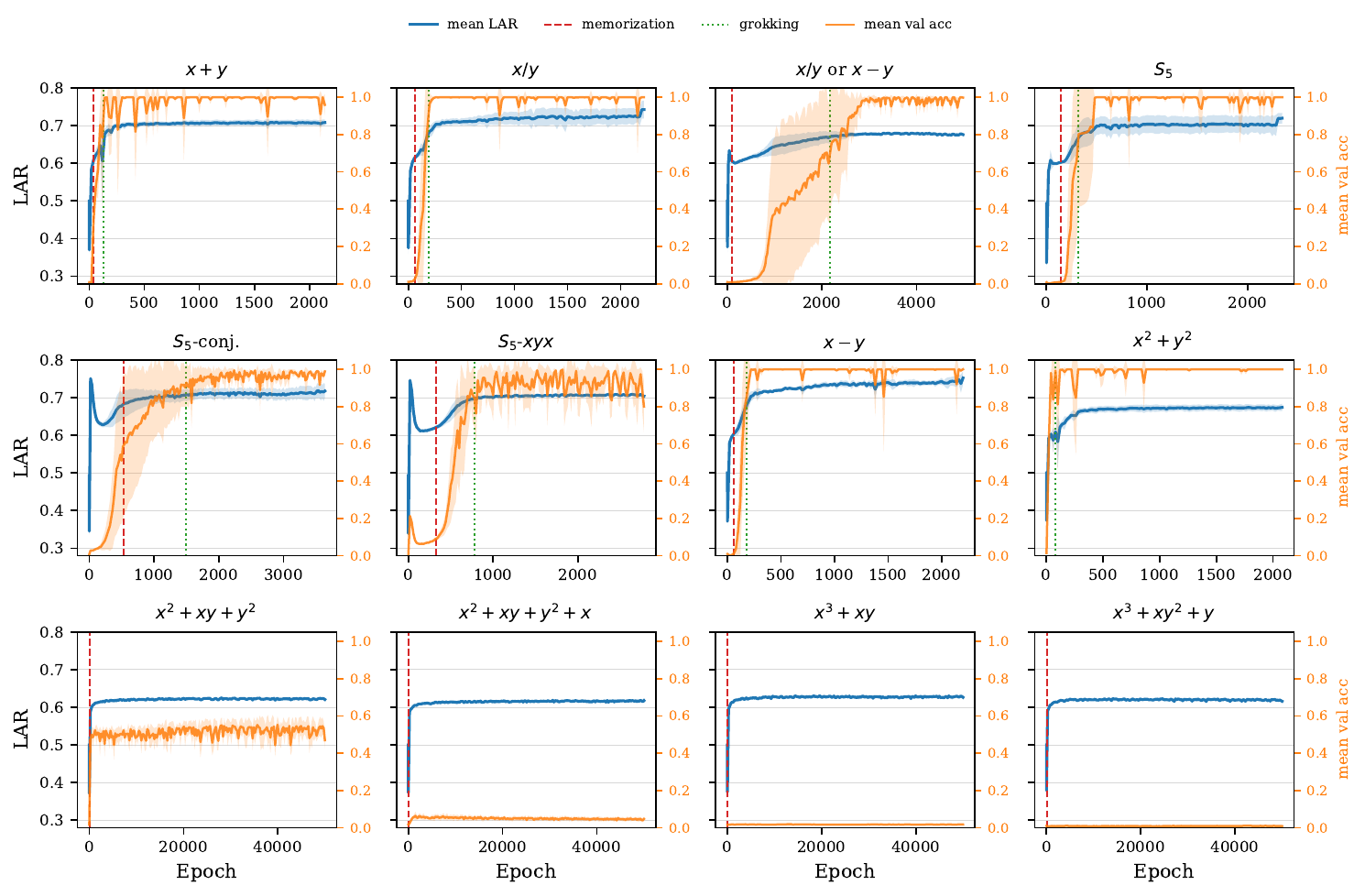}
    \caption{12 binary operation tasks.}
    \label{fig:grokking-12-tasks_full}
  \end{subfigure}
  \vspace{0.5em}
  \begin{subfigure}[t]{\columnwidth}
    \centering
    \includegraphics[width=\columnwidth]{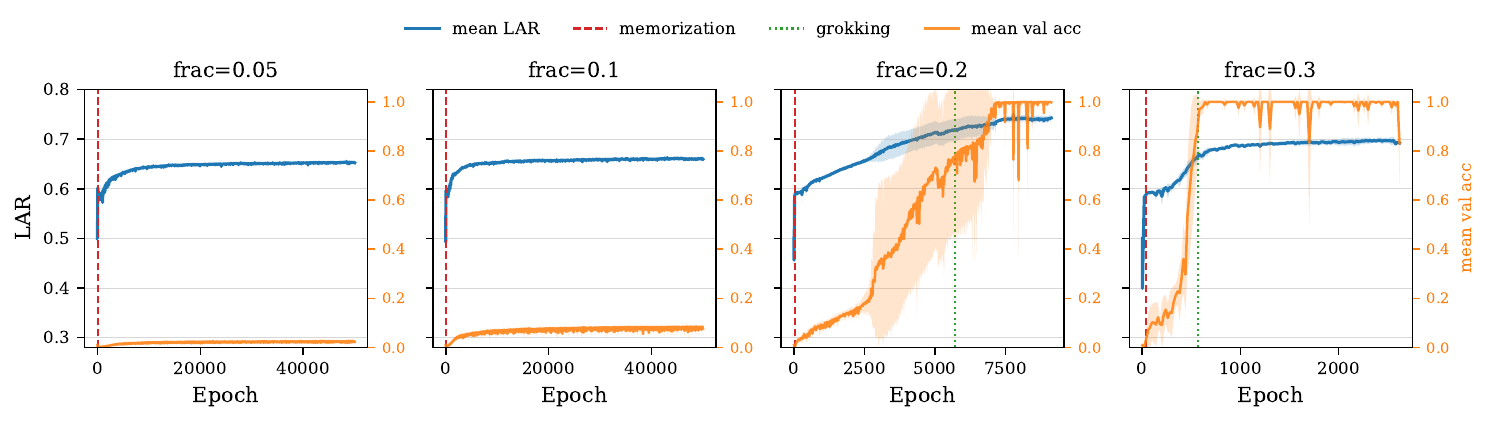}
    \caption{Modular addition with varying training dataset fraction.}
    \label{fig:grokking_dataset_fraction_full}
  \end{subfigure}
  \caption{Unembedding LAR (blue) and validation accuracy (orange) vs. epochs, with mean and ±1 std shaded bands across 5 seeds. In contrast to \cref{fig:grokking-lar-combined}, where x-axis is clipped, the full range of training epochs is shown here. The convergence of LAR under prolonged training is hence more apparent.}
  \label{fig:grokking-lar-combined-full}
\end{figure}

\end{document}